\documentclass[journal]{IEEEtran}
\ifCLASSINFOpdf
\else
\fi
\usepackage{times}
\usepackage{epsfig}
\usepackage{graphicx}
\usepackage{amsmath}
\usepackage{amssymb}
\usepackage[linesnumbered,ruled,lined]{algorithm2e}
\usepackage{graphicx} 
\usepackage{subfigure}
\usepackage{cite}
\usepackage{array}
\usepackage{amsfonts}
\usepackage{booktabs}
\usepackage{multirow}
\usepackage{tablefootnote}
\usepackage{enumitem}
\usepackage{threeparttable}
\usepackage{bm}
\usepackage[pagebackref=true,breaklinks=true,letterpaper=true,colorlinks,bookmarks=false]{hyperref}

\everymath{\displaystyle}
\setenumerate[1]{itemsep=0pt,partopsep=1pt,parsep=\parskip,topsep=1pt, leftmargin=13pt}
\setitemize[1]{itemsep=0pt,partopsep=1pt,parsep=\parskip,topsep=1pt, leftmargin=13pt}
\setdescription{itemsep=0pt,partopsep=1pt,parsep=\parskip,topsep=1pt, leftmargin=13pt}

\makeatletter
\def\UrlAlphabet{%
      \do\a\do\b\do\c\do\d\do\e\do\f\do\g\do\h\do\i\do\j%
      \do\k\do\l\do\m\do\n\do\o\do\p\do\q\do\r\do\s\do\t%
      \do\u\do\v\do\w\do\x\do\y\do\z\do\A\do\B\do\C\do\D%
      \do\E\do\F\do\G\do\H\do\I\do\J\do\K\do\L\do\M\do\N%
      \do\O\do\P\do\Q\do\R\do\S\do\T\do\U\do\V\do\W\do\X%
      \do\Y\do\Z}
\def\UrlDigits{\do\1\do\2\do\3\do\4\do\5\do\6\do\7\do\8\do\9\do\0}
\g@addto@macro{\UrlBreaks}{\UrlOrds}
\g@addto@macro{\UrlBreaks}{\UrlAlphabet}
\g@addto@macro{\UrlBreaks}{\UrlDigits}
\makeatother

\begin{document}
%
\title{Pixel-Stega: Generative Image Steganography Based on Autoregressive Models}
%
%
%

\author{Siyu~Zhang,
        Zhongliang~Yang,
        Haoqin~Tu,
        Jinshuai~Yang,
        and~Yongfeng~Huang
\thanks{S. Zhang, Z. Yang, J. Yang and Y. Huang were with the Department of Electronic Engineering, Tsinghua University, Beijing,
 China. E-mail: yangzl15@tsinghua.org.cn.}
\thanks{H. Tu was with the School of Cyber Security, University of Chinese Academy of Sciences, Beijing.}
}

\maketitle

\begin{abstract}
In this letter, we explored generative image steganography based on autoregressive models. We proposed \emph{Pixel-Stega}, which implements pixel-level information hiding with autoregressive models and arithmetic coding algorithm. Firstly, one of the autoregressive models, PixelCNN++, is utilized to produce explicit conditional probability distribution of each pixel. Secondly, secret messages are encoded to the selection of pixels through steganographic sampling (stegosampling) based on arithmetic coding. We carried out qualitative and quantitative assessment on gray-scale and colour image datasets. Experimental results show that Pixel-Stega is able to embed secret messages adaptively according to the entropy of the pixels to achieve both high embedding capacity (up to 4.3 bpp) and nearly perfect imperceptibility (about 50\% detection accuracy).\footnote{Our code is available at \url{https://github.com/Mhzzzzz/Pixel-Stega}.}
\end{abstract}


%
\IEEEpeerreviewmaketitle

%
%
%
%

 





\section{Introduction}
\label{section introduction}
\IEEEPARstart{W}ith the rapid development of surveillance technology and the increasingly severe threat to information security \cite{greenwald2013xkeyscore, landau2013making}, the protection of private communication has aroused more and more attention in recent years \cite{ventures2017cybersecurity}. Steganography is one of the most predominant techniques against eavesdropping \cite{mishra2015review}, which refers to the problem of sending secret messages embedded in seemingly innocuous carriers  over a public channel so that an eavesdropper cannot even detect the presence of the hidden messages.

Digital image is the most widely investigated carrier for steganography due to its large redundancy and frequent use in cyberspace \cite{bhattacharyya2011survey, morkel2005overview}. Traditional image steganographic methods mainly embed secret messages by slightly modifying some insensitive features of the image, either in spatial domain \cite{pevny2010using, holub2012designing, holub2014universal, chen2016defining} or in transform domain \cite{westfeld2001f5, fridrich2007statistically, filler2011minimizing, chen2018defining, guo2015using, cogranne2020steganography}, while minimizing an heuristically defined distortion function. There are also some methods incorporating with automatic distortion functions based on deep learning approaches like Generative Adversarial Networks (GANs) \cite{tang2017automatic, yang2019embedding, tang2019cnn, tang2020automatic, bernard2020explicit}. 

Recent advances in the field of image generation have promoted the development of another paradigm, generative image steganography, which directly generates realistic-looking steganographic images (stego-images) according to the secret messages \cite{fridrich2009steganography}. It is more flexible but also challenging. In preliminary studies, researchers tried unnatural image synthesis for information hiding, such as texture \cite{wei1999deterministic, otori2007data, otori2009texture, wu2014steganography, xu2015hidden, pan2016steganography, qian2017robust} and fingerprint \cite{zhao2012fingerprint, li2018toward}. However, such kind of images has limited scope of application.

To generate more generic stego-images, some researchers realized generative image steganography by encoding secret messages into the inputs of generative models  \cite{hu2018novel, li2020generative, chen2018provably, arifianto2020edgan, jiang2020new, liu2017coverless, zhang2020generative, luo2020coverless}. For example, Hu {\it et al.} \cite{hu2018novel} embedded secret messages into a latent vector $z$, and transformed it into realistic image with the generator of GAN. They trained an additional extractor network for information recovery. Li {\it et al.} \cite{li2020generative} improved the method by jointly training the generator and the extractor. Chen {\it et al.} \cite{chen2018provably} used a similar strategy with the encoder of a variational autoencoder (VAE). Liu {\it et al.} \cite{liu2017coverless} encoded secret messages to the selection of class labels of an Auxiliary Classifier Generative Adversarial Network (ACGAN). However, these methods generally suffer from low embedding capacity, as well as inaccurate information extraction.

\begin{figure*}[t]
  \centering
  \includegraphics[width=\textwidth]{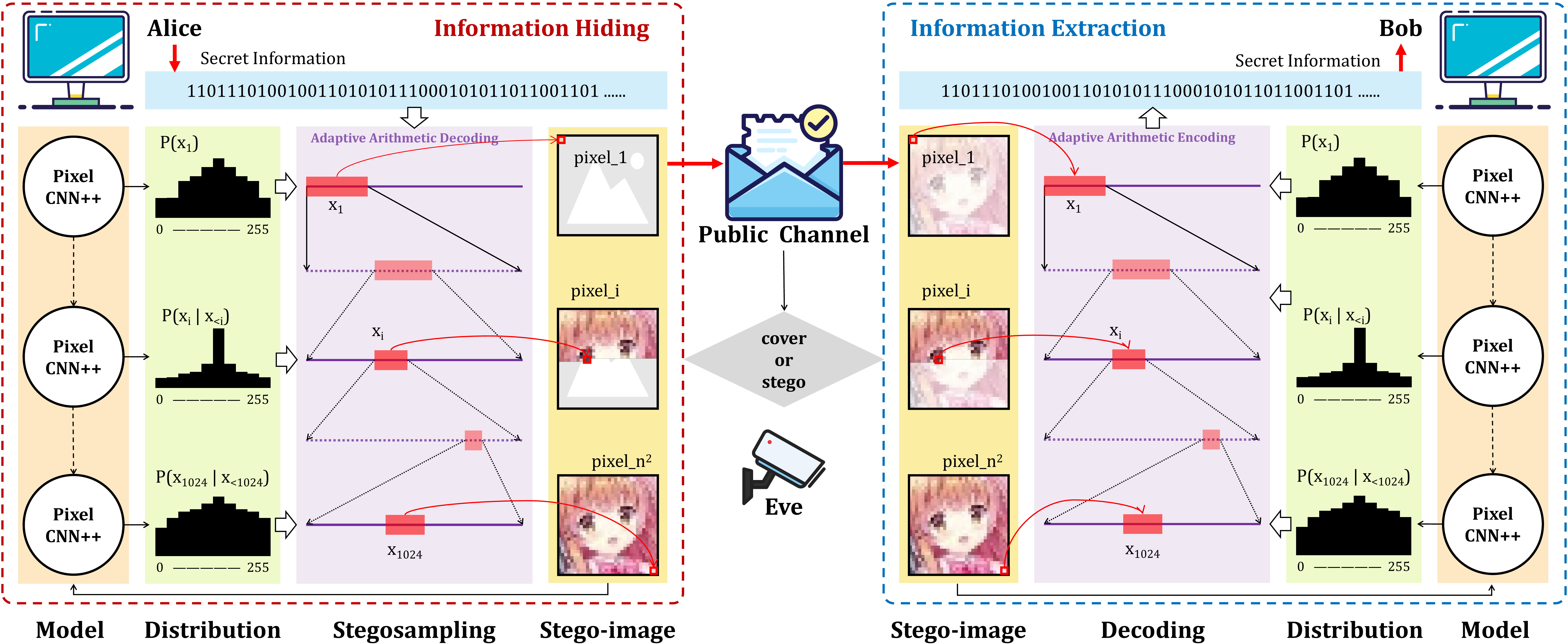}
  \caption{The overall framework of Pixel-Stega. PixelCNN++ produces explicit conditional probability distribution autoregressively. At Alice's end, Pixel-Stega encodes 256 pixel values by stegosampling based on arithmetic coding, and then selects the pixel value corresponding to the secret message. The stego-image is then transmitted through a lossless public channel monitored by Eve. At Bob's end, an inverse process is executed to extract the secret message.}
  \label{pixel-stega figure}
\end{figure*}

\begin{figure}[htbp]
  \centering \includegraphics[width=2.2in]{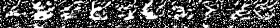}
  \caption{Stego-images of MNIST generated by the method proposed in \cite{yang2018provably}.}
  \label{rejection mnist figure}
\end{figure}

To circumvent these difficulties, another line of thought tried to embed secret messages via the outputs of generative models instead of inputs. 
On condition that the output probability distribution of an image is known, the embedding and extraction of generative image steganography is exactly the equivalent problem of source decoding and encoding, respectively. Specifically, in steganographic embedding or source decoding, secret messages/codes are transformed into images, while in steganographic extraction or source encoding, images are transformed into secret messages/codes. Therefore, a possible way to achieve generative image steganography is to deploy steganographic sampling (stegosampling) strategy based on lossless source coding algorithm on explicit generative models, such as autoregressive models.
On the one hand, lossless source coding enables the absolutely accurate extraction. On the other hand, the secret message is able to be encoded into 256 sampling values of all pixels in an image, which enables a considerable embedding capacity as long as the conditional distribution has a large entropy.
There have been attempts in generative image steganography based on autoregressive models. Yang {\it et al.} \cite{yang2018provably} proposed a rejection-sampling-based strategy to embed secret messages into the Least Significant Bit (LSB) of each pixel during generation. However, this method is not adaptive to sharp conditional distribution so that it may fail to generate meaningful stego-images, as demonstrated in Figure \ref{rejection mnist figure}.

In this letter, we further explored generative image steganography based on autoregressive models. We are committed to improving the stegosampling algorithm to pursue both high embedding capacity and high imperceptibility. We proposed \emph{Pixel-Stega}, which employed the same autoregressive model as Yang {\it et al.} \cite{yang2018provably}, but adopted arithmetic coding algorithm (which has high coding rate) to construct the stegosampling strategy. Through quantitative analysis, we observed that Pixel-Stega can embed up to 4.3 bits information per pixel in average (which is approximately the entropy of the pixels) and achieved about 50\% detection accuracy. We also noticed that Pixel-Stega embeds secret messages proportional to the entropy of pixels, which ensures that the explicit probability distribution is fully utilized and the proposed method is adaptable to all kinds of distribution (no matter sharp or flat).

\section{Pixel-Stega Methodology}
\label{section pixel-stega methodology}

\subsection{Overview}
\label{subsection overview}
Figure \ref{pixel-stega figure} illustrates the overall framework. It is supposed that the sender Alice wants to send a secret message $m \sim \mbox{Uniform}(0,1)^L$ to the receiver Bob, while the channel is monitored by an eavesdropper Eve \cite{simmons1984prisoners}. In Pixel-Stega, they share the same autoregressive model like PixelCNN++ for information embedding and extraction. Pixel-Stega works in two steps. Firstly, image pixels are modeled as a sequence from top left to bottom right. PixelCNN++ produces the conditional probability distribution $p(x_i|x_{<i})$ of the current ($i$-th) pixel $x_i$, which is a discrete distribution over the 256 pixel values. For colour images in RGB format, we consider $x_i$ as the pixel in R channel. Secondly, stegosampling based on arithmetic coding is applied. We regard the pixel sequence as symbols and the secret message as codes. At Alice's end, Pixel-Stega takes the explicit distribution $p(x_i|x_{<i})$ and the secret message $m$ as input and output the selected value $v$ of the current pixel ($v \in \{0,1,...,255\}$)  by arithmetic decoding. At Bob's end, the secret message $m$ is extracted by taking the distribution $p(x_i|x_{<i})$ and the pixel value $v$ as the input of arithmetic encoding algorithm.

\begin{figure*}[htbp]
  \centering
  \subfigure[Stego-images of MNIST]{\begin{minipage}[t]{0.32\linewidth} \centering \includegraphics[width=2.2in]{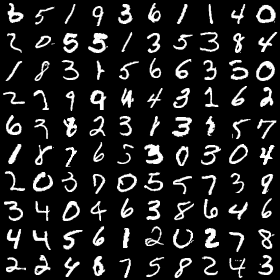} \end{minipage}\label{MNIST example subfigure}}
  \subfigure[Stego-images of Frey Faces]{\begin{minipage}[t]{0.32\linewidth} \centering \includegraphics[width=2.2in]{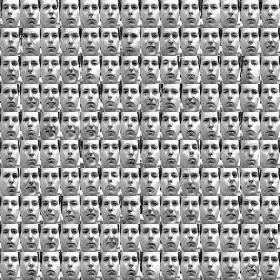} \end{minipage}}
  \subfigure[Stego-images of CIFAR-10]{\begin{minipage}[t]{0.32\linewidth} \centering \includegraphics[width=2.2in]{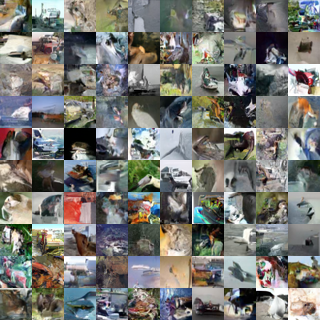} \end{minipage}}
  \centering
  \caption{Stego-images generated by Pixel-Stega.}
  \label{example figure}
\end{figure*}

\subsection{Stegosampling Based on Arithmetic Coding}
\label{subsection stegosampling based on arithmetic coding}

\begin{figure}[h]
  \centering
  \includegraphics[width=\columnwidth]{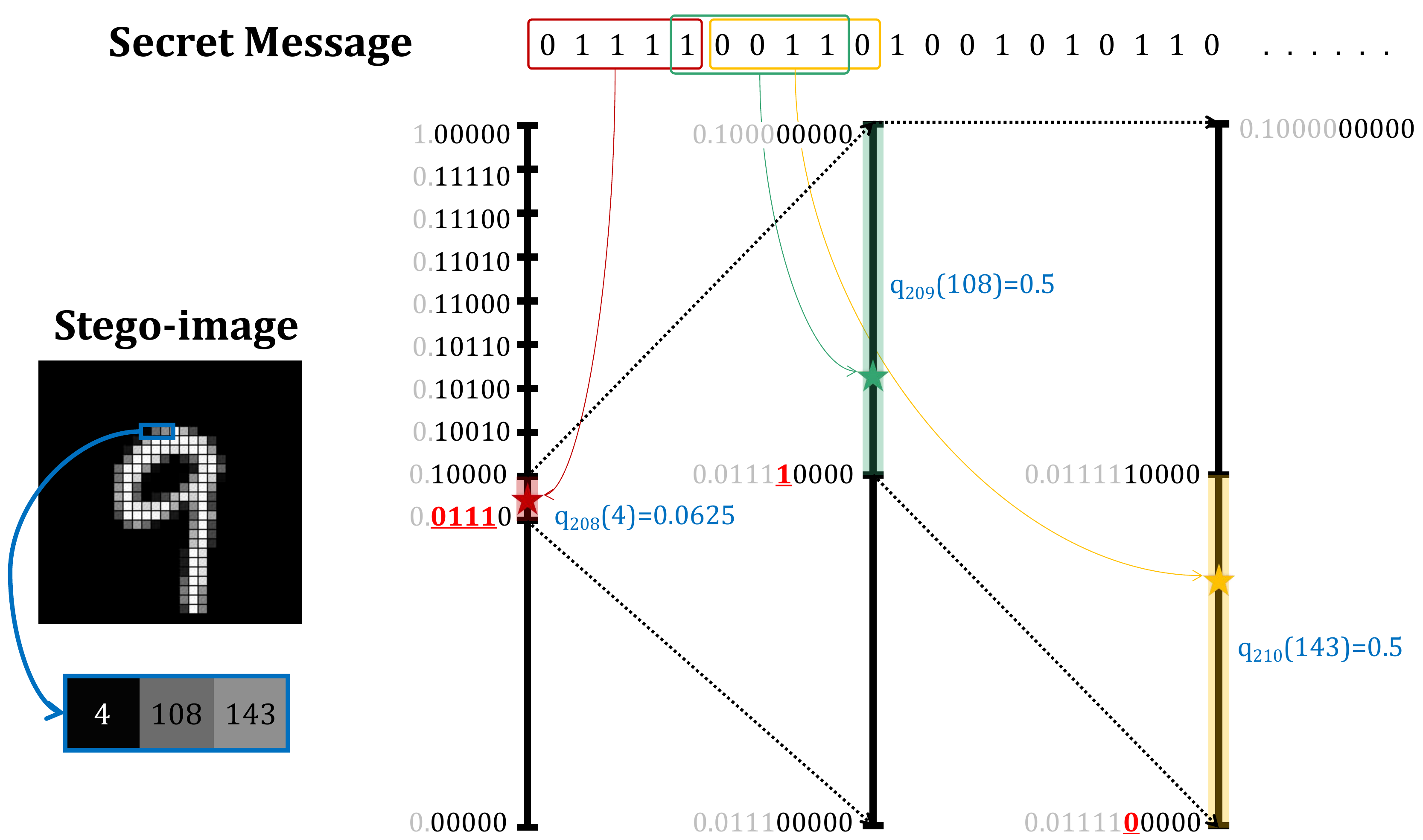}
  \caption{Details in stegosampling based on arithmetic coding ($prc=5$). It is considered as an adaptive arithmetic decoding process, where the secret message plays a role as arithmetic code and pixel sequence is the original symbol flow.}
  \label{stegosampling figure}
\end{figure}

Arithmetic coding is a kind of entropy coding used in lossless data compression, which encodes a symbol flow by creating a code which represents a fractional value on a real number between 0 and 1 \cite{witten1987arithmetic}. On each symbol recursion, an interval $[l_i,u_i)$ (which is initially set to $[0,1)$) is successively partioned according to the cumulative distribution of the symbol, and narrowed down by taking the sub-interval containing the symbol value as the new interval $[l_{i+1},u_{i+1})$. Following Zigler {\it et al.} \cite{ziegler2019neural}, we adopted a fixed-precision implementation of arithmetic coding \cite{rubin1979arithmetic}.


\subsubsection{Steganographic Embedding}
In practice, the required precision of digits grows as encoding continues. To mitigate this issue, fixed-precision implementation of arithmetic coding uses $prc$-bit registers to store the mantissa of the code and the boundaries of the interval. On this basis, Pixel-Stega generates each pixel $x_i$ by three steps. Firstly, the conditional probability distribution $p(x_i|x_{<i})$ is sorted and the current interval $[l_i,u_i)$ is partioned into sub-intervals proportional to its cumulative distribution $c(x_i|x_{<i})$. This procedure can be seen as a quantization of the distributions and  we obtain the quantized probability distribution $q(x_i|x_{<i})$. The quantization interval is the current minimum decimal that the registers can represent. Secondly, we determine pixel $x_i$ to be the value that its sub-interval contains the real number represented by the secret message. We can locate the sub-interval referring to the register storing the secret message. Finally, the same prefix of all numbers in the new interval is the newly embedded secret message. No matter how the embedding procedure continues later, this prefix will not be changed anymore. Thus, registers storing boundries shift left to overflow the prefix, and the register storing the secret message (more like a sliding window) also shifts left (the sliding window moves right) by the same number of digits.

An example is illustrated in Figure \ref{stegosampling figure}. At the first embedding step ($i=208$), the interval still remains initialized as $[0.00000,1.00000)$. The interval is partioned in proportion to $q(x_{208}|x_{<208})$. According to the first $prc$-bit secret message 01111, the sub-interval $[0.01110,0.10000)$ with a pixel value of 4 is selected as the next interval. Since the \emph{open} boundary 0.10000 is not available, all numbers in the sub-interval starts with the prefix 0.0111. As a result, it is considered that the first 4 bit secret message is encoded to the 208-th pixel. The registers storing boundaries and the secret message shift left by 4 bits to overflow the prefix. This process is repeated until the stego-image is completed.

\subsubsection{Steganographic Extraction}
The extraction procedure is quite an inverse process of the embedding. Firstly, the same operation is conducted to obtain the sorted distributions and partion the interval. Secondly, the current pixel $x_i$ locates the next interval. Finally, we extract the same prefix of numbers in the next interval as the secret message. Registers shift left by the number of digits.

\subsection{Imperceptibility}
\label{subsection imperceptibility}
As is often assumed, the secret message $m$ consists of uniformly distributed random bits, hence the pixel is equivalent to be sampled from the quantized distribution $q(x_i|x_{<i})$, which reveals the advantage of Pixel-Stega in two respects. On the one hand, it is able to make the most of the information provided by the output distribution $p(x_i|x_{<i})$ thus provides a strong guarantee for embedding capacity. The average amount of information that can be embedded at each time step of Pixel-Stega is given by the entropy of the quantized distribution $H(q(x_i|x_{<i}))$, and affected by $H(p(x_i|x_{<i}))$. This corresponds to the characteristic of high compression rate (close to entropy) of arithmetic coding \cite{rissanen1979arithmetic}. On the other hand, it ensures high imperceptibility as each pixel is determined adaptively on the basis of $q(x_i|x_{<i})$. Distortion between the equivalent distribution $q(x_i|x_{<i})$ and the original distribution $p(x_i|x_{<i})$ is slight enough. If the impact of truncation, quantization and blocking of secret message is ignored ($prc$ tends to infinity and the secret message $m$ has infinite length), $q(x_i|x_{<i})$ is actually identical to $p(x_i|x_{<i})$ and it is able to achieve the optimal imperceptibility. Arithmetic coding is widely applied in generative linguistic and audio steganography due to its advantage in embedding capacity and imperceptibility \cite{chen2018provably, yang2019linguistic, ziegler2019neural}. But as far as we know, there hasn't been any research about generative image steganography with arithmetic-coding-based stegosampling.

\begin{figure*}[htbp]
  \centering
  \subfigure[Stego-images]{
  \begin{minipage}[t]{0.095\textwidth} \centering \includegraphics[width=0.65in]{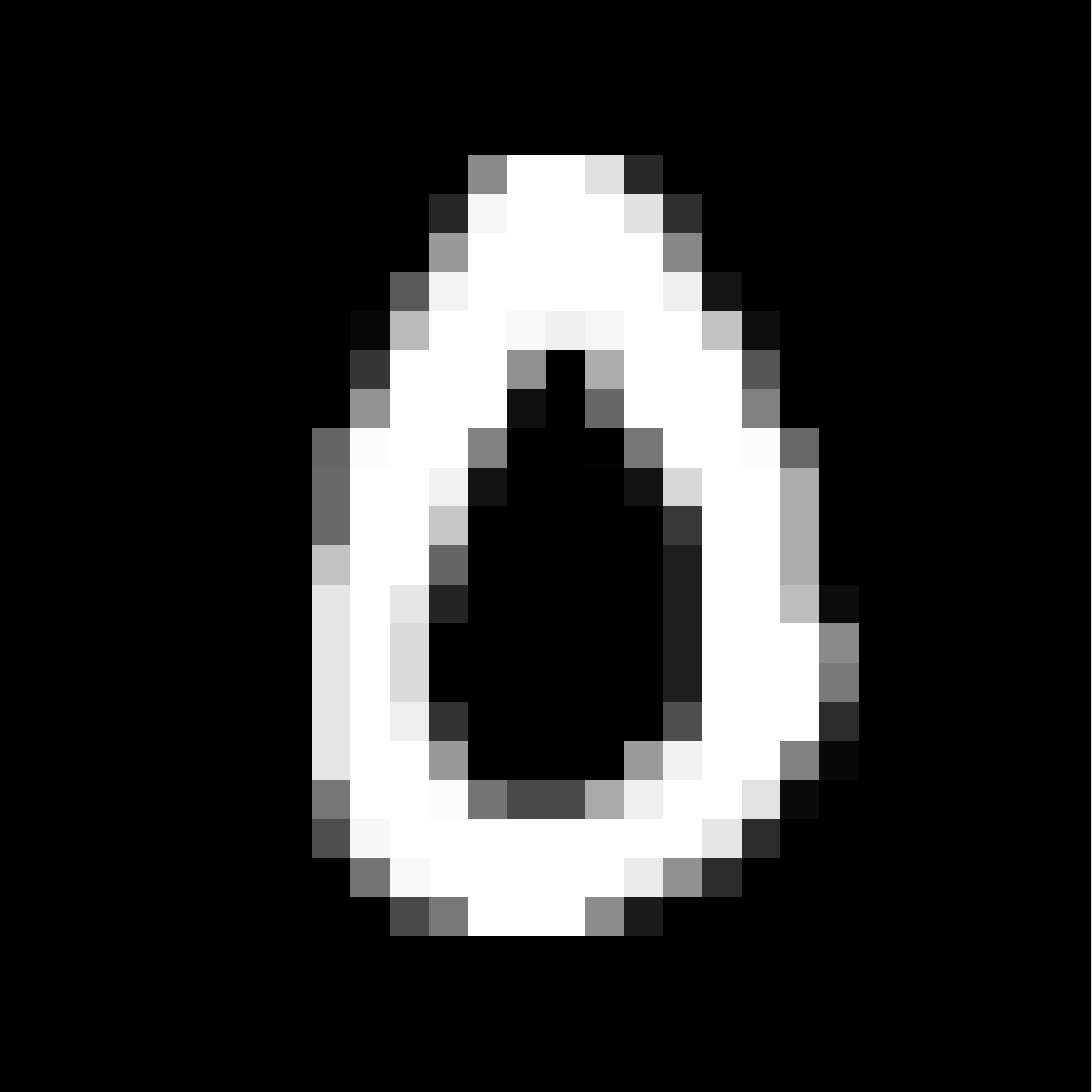} \end{minipage}
  \begin{minipage}[t]{0.095\textwidth} \centering \includegraphics[width=0.65in]{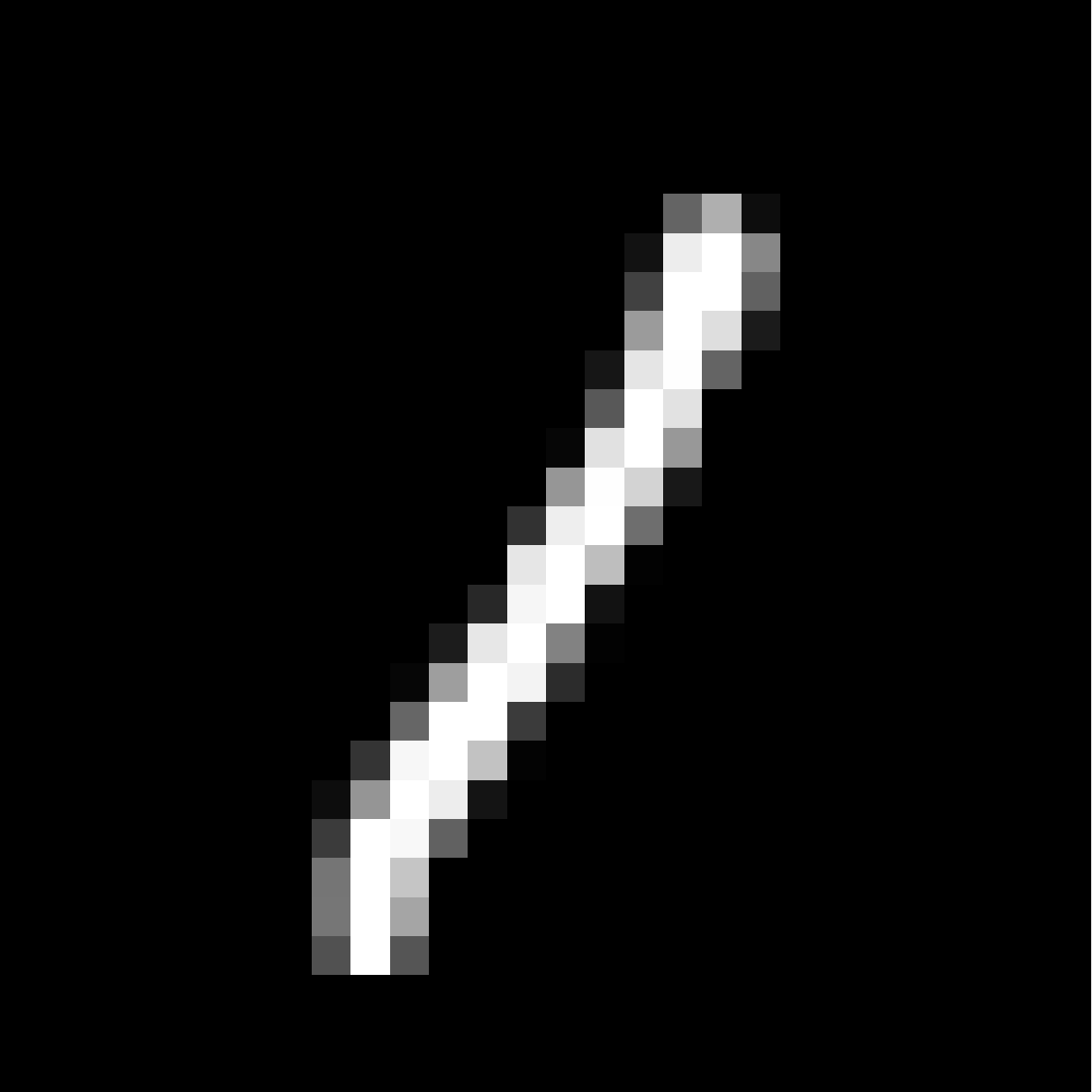} \end{minipage}
  \begin{minipage}[t]{0.095\textwidth} \centering \includegraphics[width=0.65in]{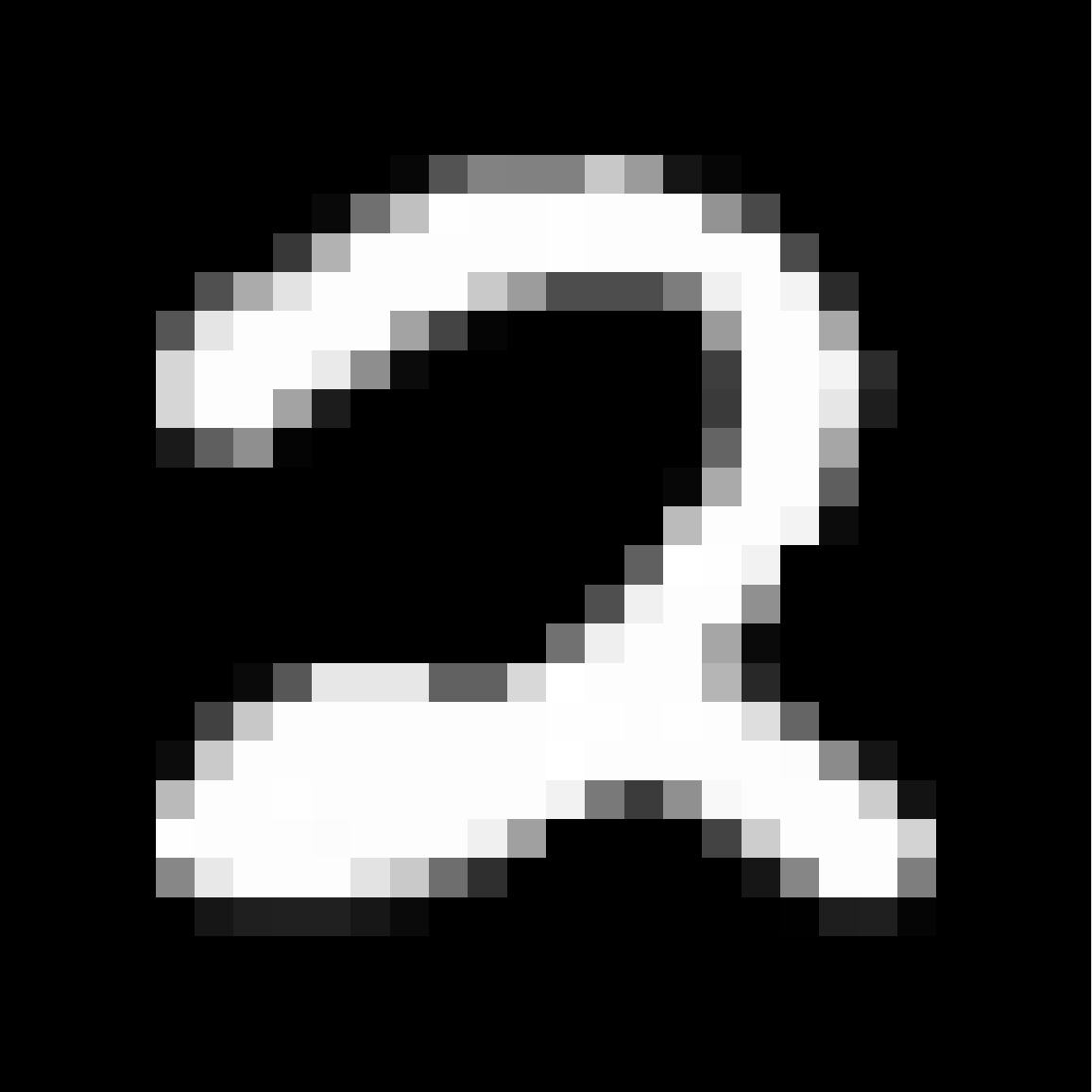} \end{minipage}
  \begin{minipage}[t]{0.095\textwidth} \centering \includegraphics[width=0.65in]{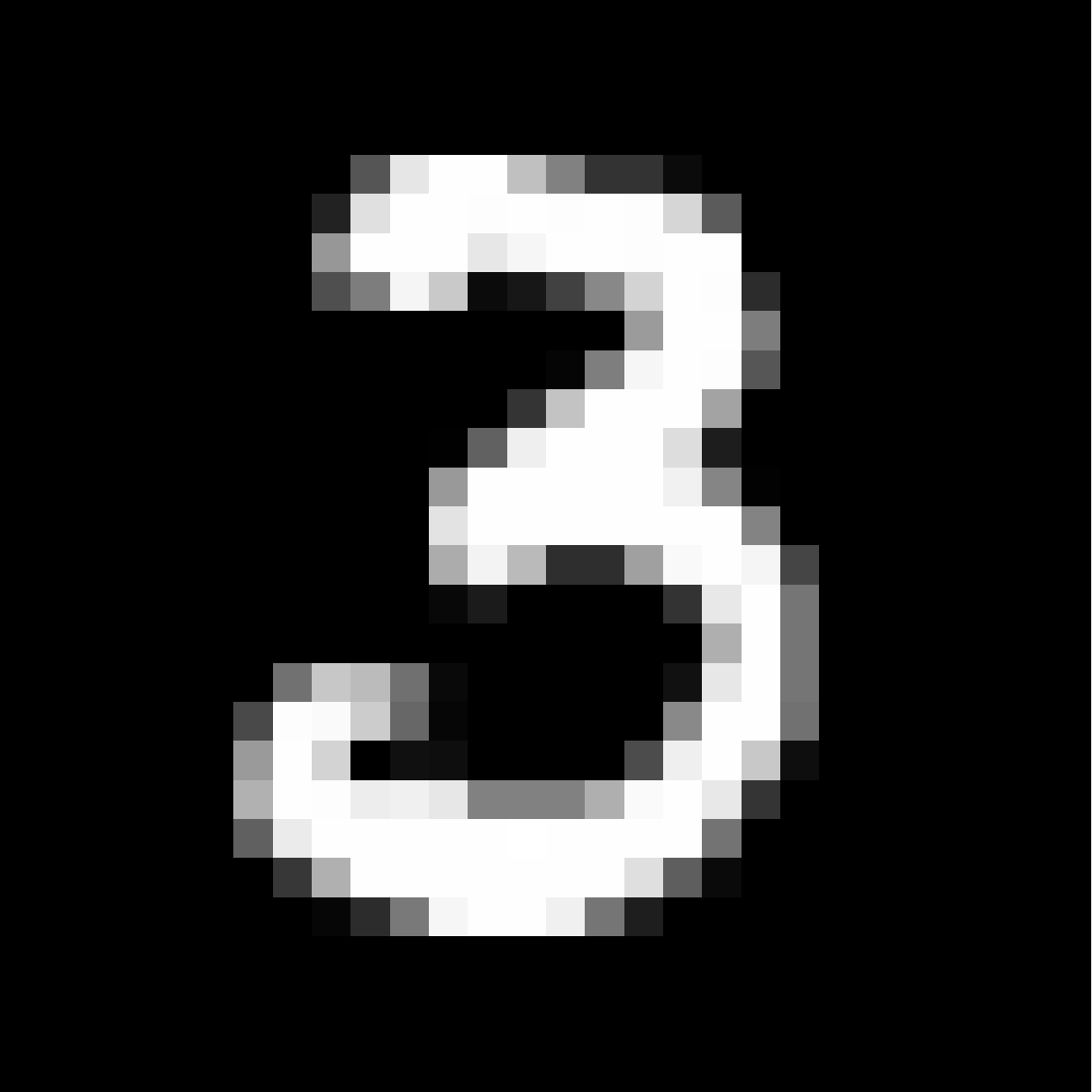} \end{minipage}
  \begin{minipage}[t]{0.095\textwidth} \centering \includegraphics[width=0.65in]{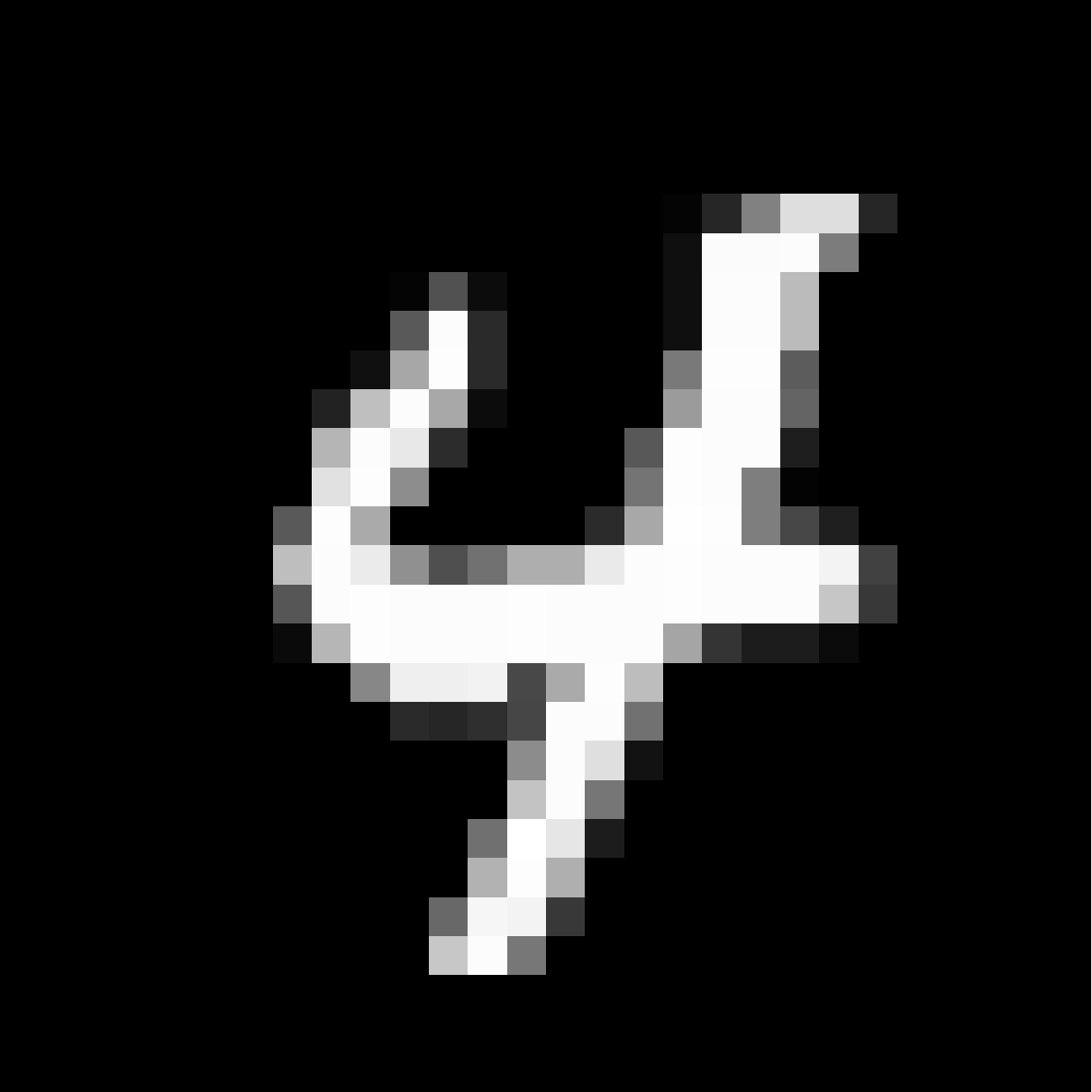} \end{minipage}
  \begin{minipage}[t]{0.095\textwidth} \centering \includegraphics[width=0.65in]{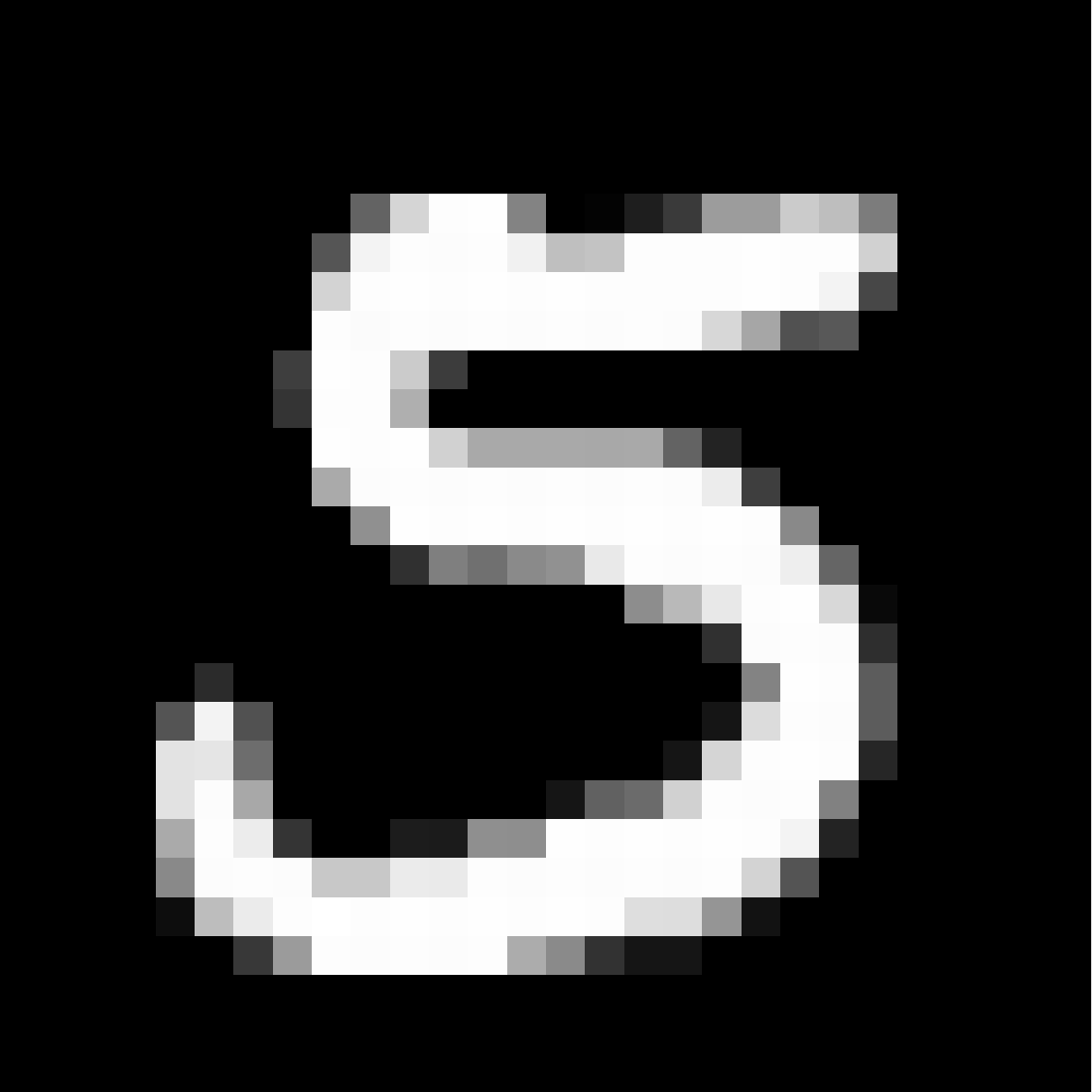} \end{minipage}
  \begin{minipage}[t]{0.095\textwidth} \centering \includegraphics[width=0.65in]{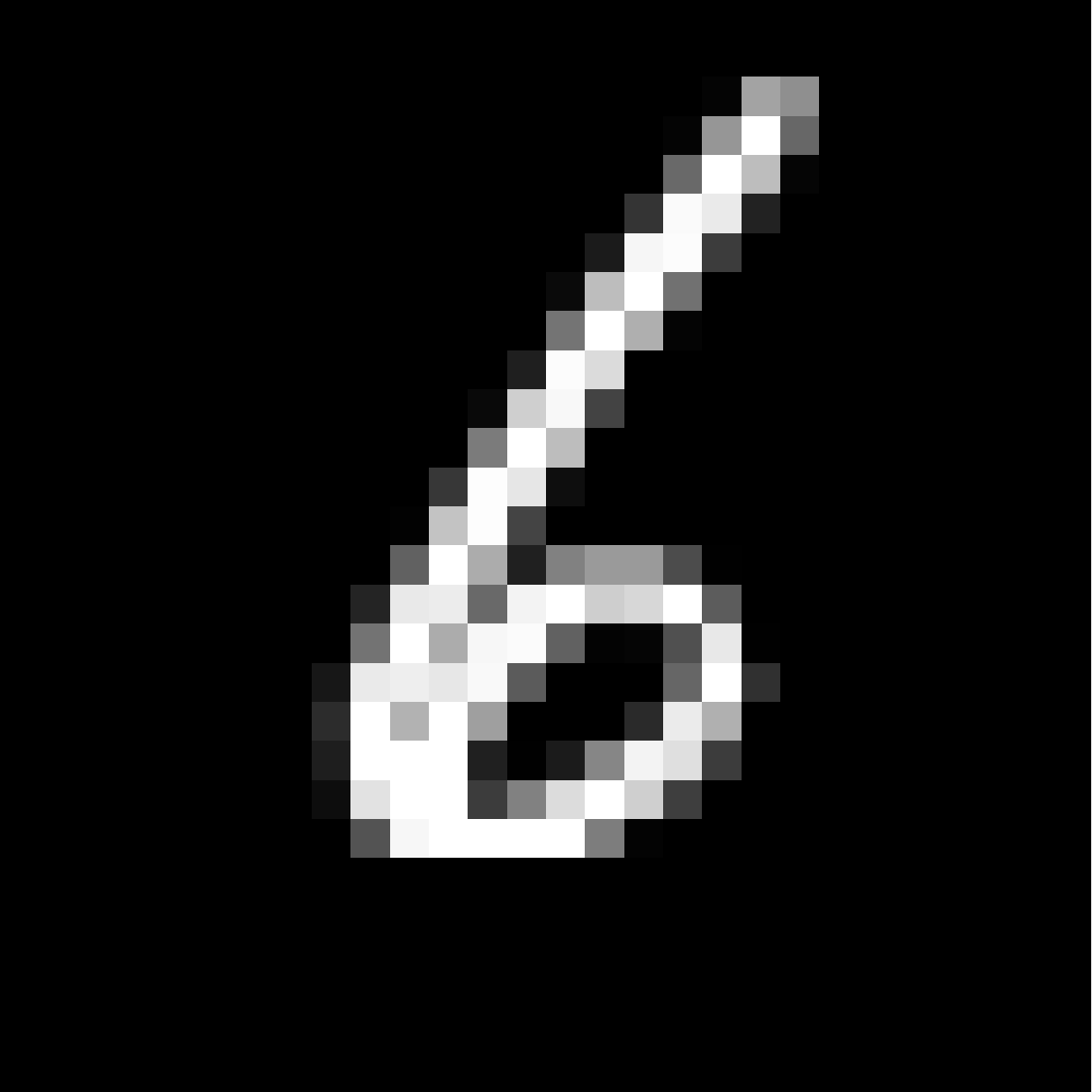} \end{minipage}
  \begin{minipage}[t]{0.095\textwidth} \centering \includegraphics[width=0.65in]{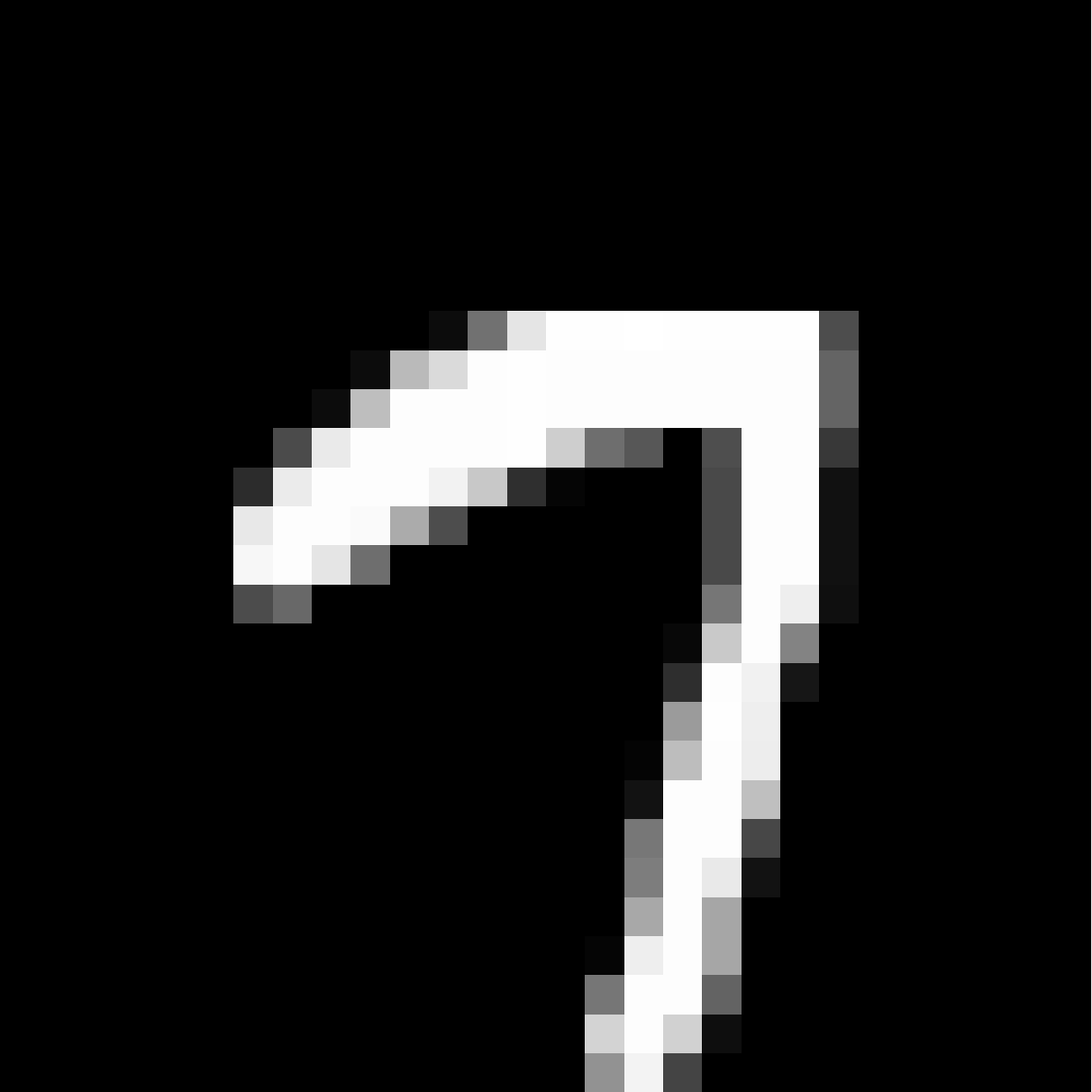} \end{minipage}
  \begin{minipage}[t]{0.095\textwidth} \centering \includegraphics[width=0.65in]{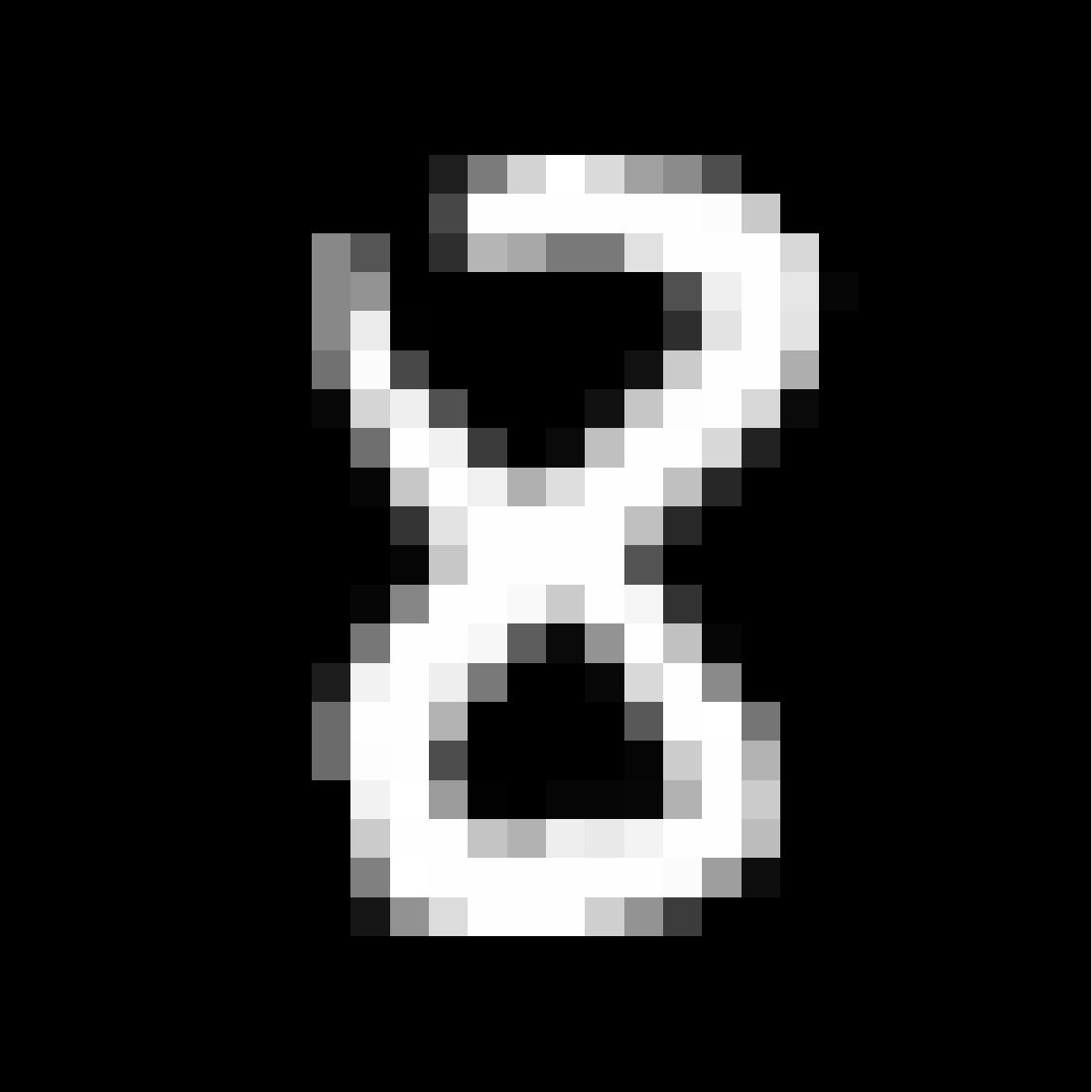} \end{minipage}
  \begin{minipage}[t]{0.095\textwidth} \centering \includegraphics[width=0.65in]{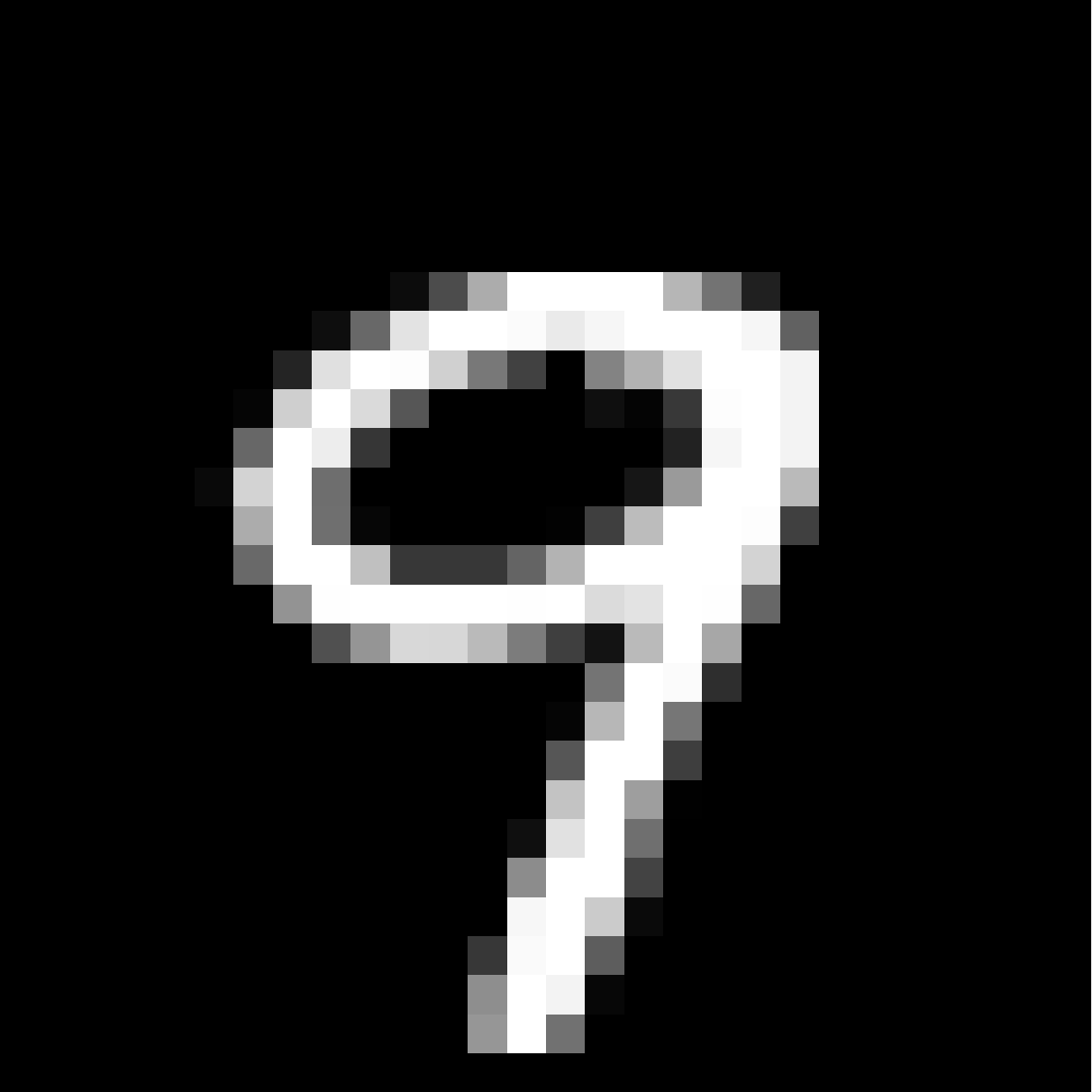} \end{minipage}
  \label{entropy subfigure}}
  \centering
  \subfigure[Entropy]{
  \begin{minipage}[t]{0.095\textwidth} \centering \includegraphics[width=0.65in]{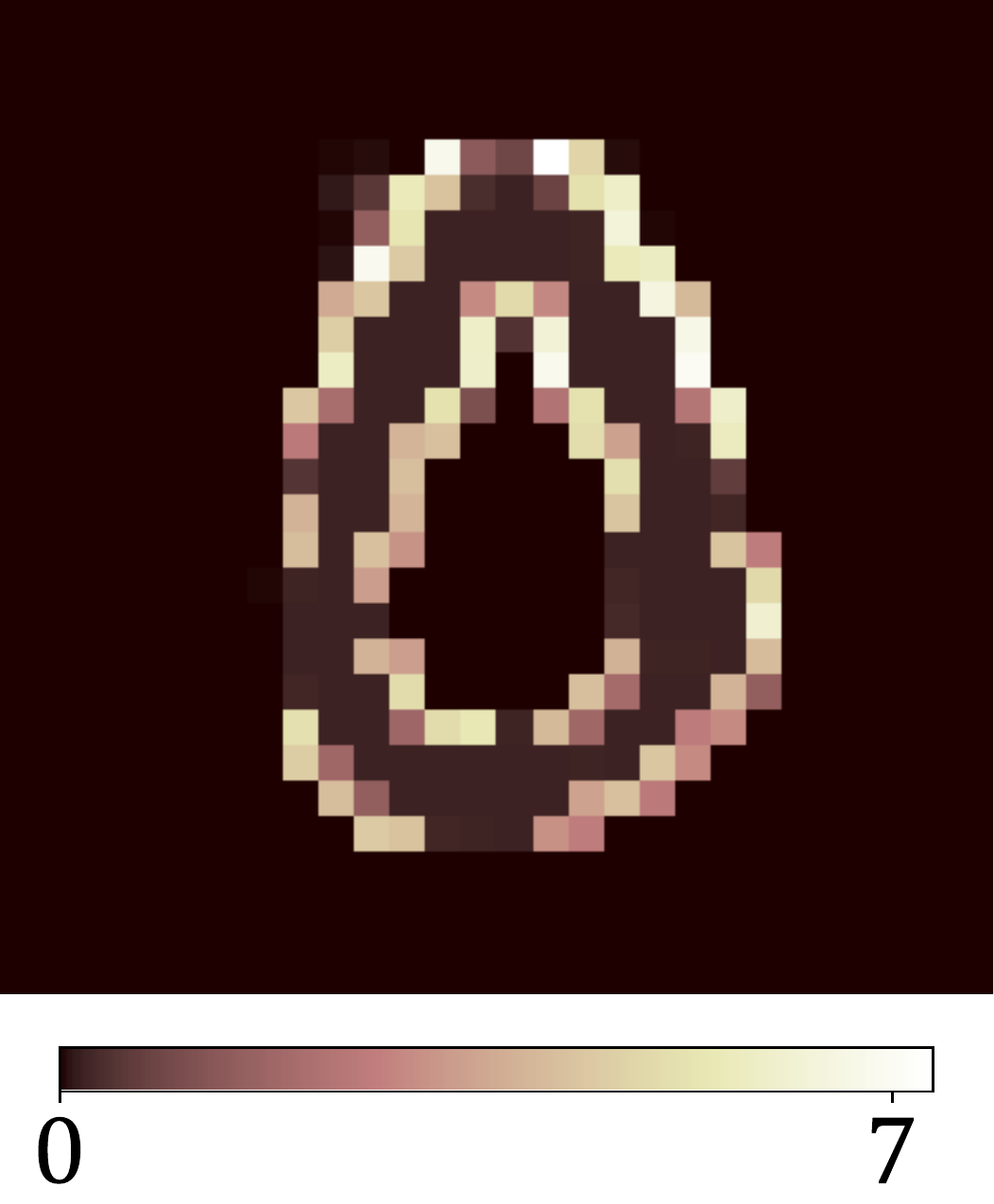} \end{minipage}
  \begin{minipage}[t]{0.095\textwidth} \centering \includegraphics[width=0.65in]{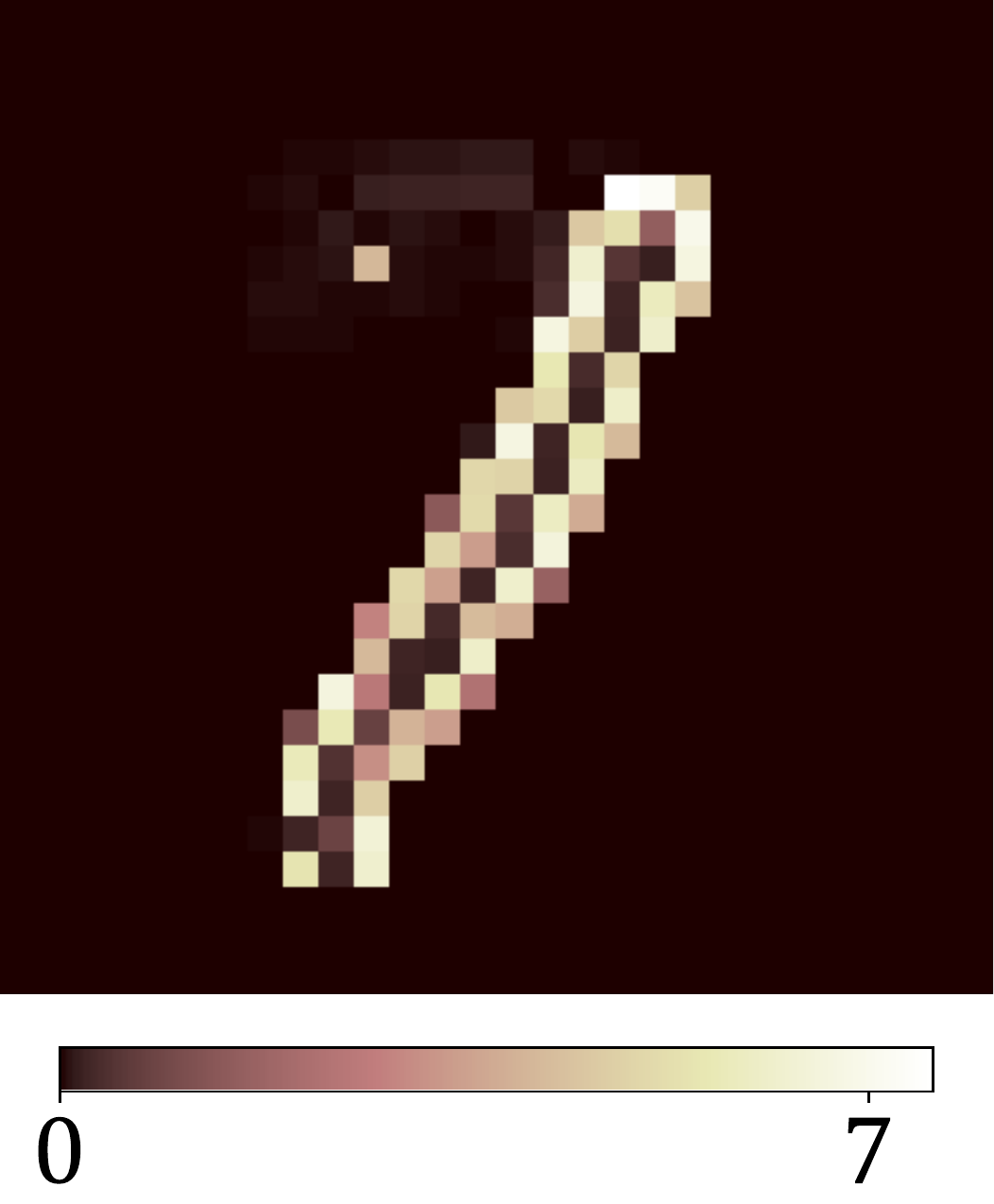} \end{minipage}
  \begin{minipage}[t]{0.095\textwidth} \centering \includegraphics[width=0.65in]{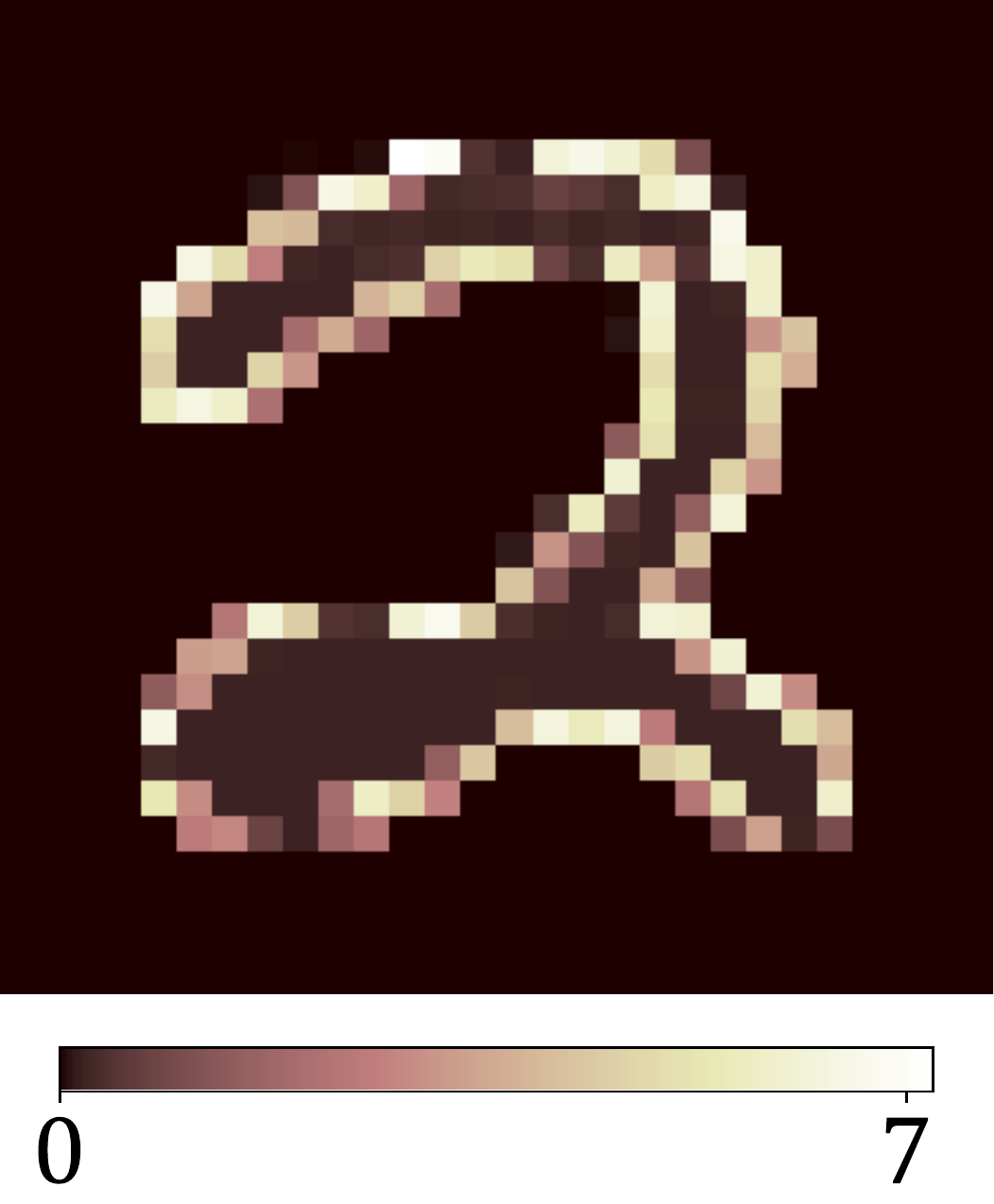} \end{minipage}
  \begin{minipage}[t]{0.095\textwidth} \centering \includegraphics[width=0.65in]{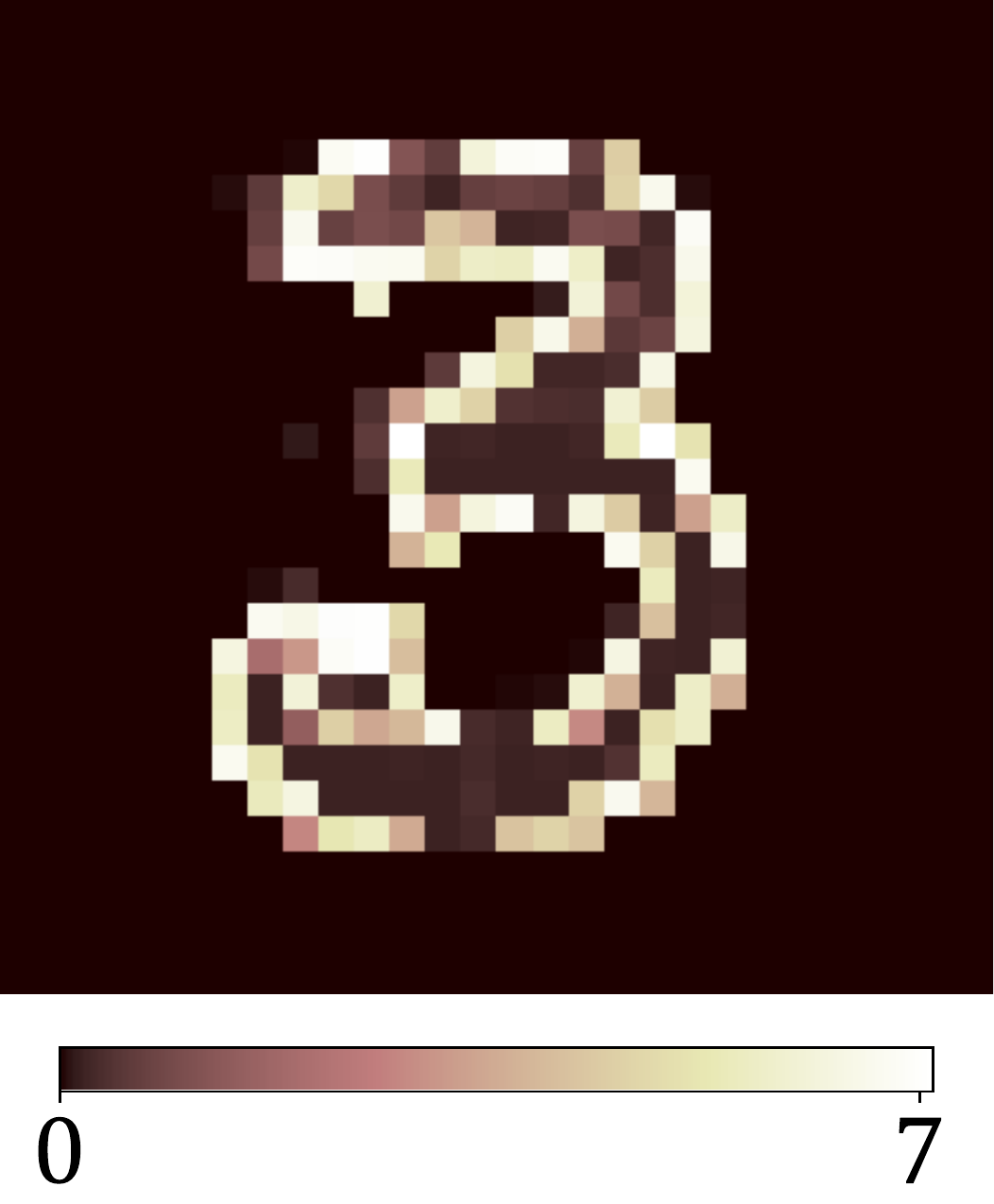} \end{minipage}
  \begin{minipage}[t]{0.095\textwidth} \centering \includegraphics[width=0.65in]{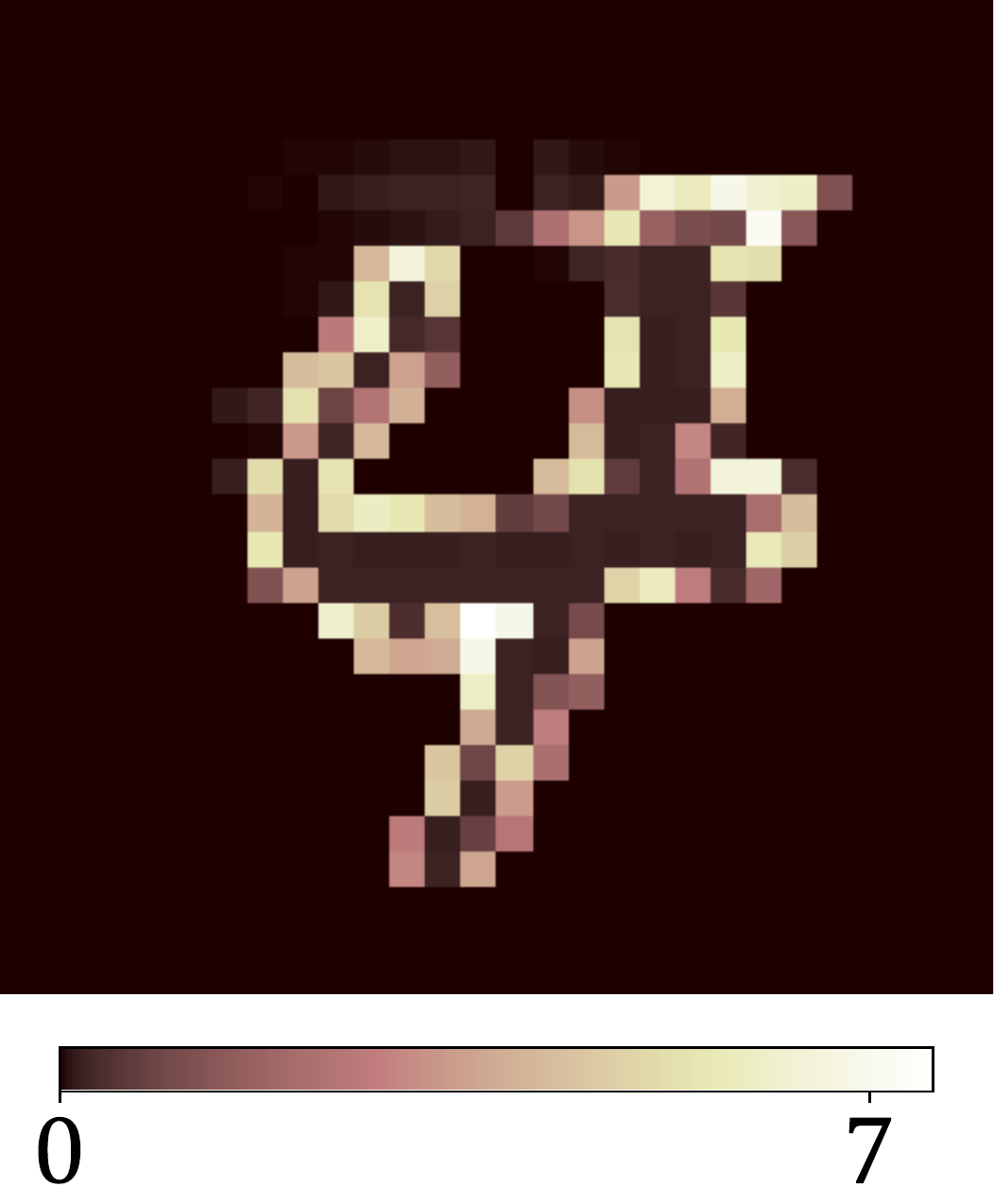} \end{minipage}
  \begin{minipage}[t]{0.095\textwidth} \centering \includegraphics[width=0.65in]{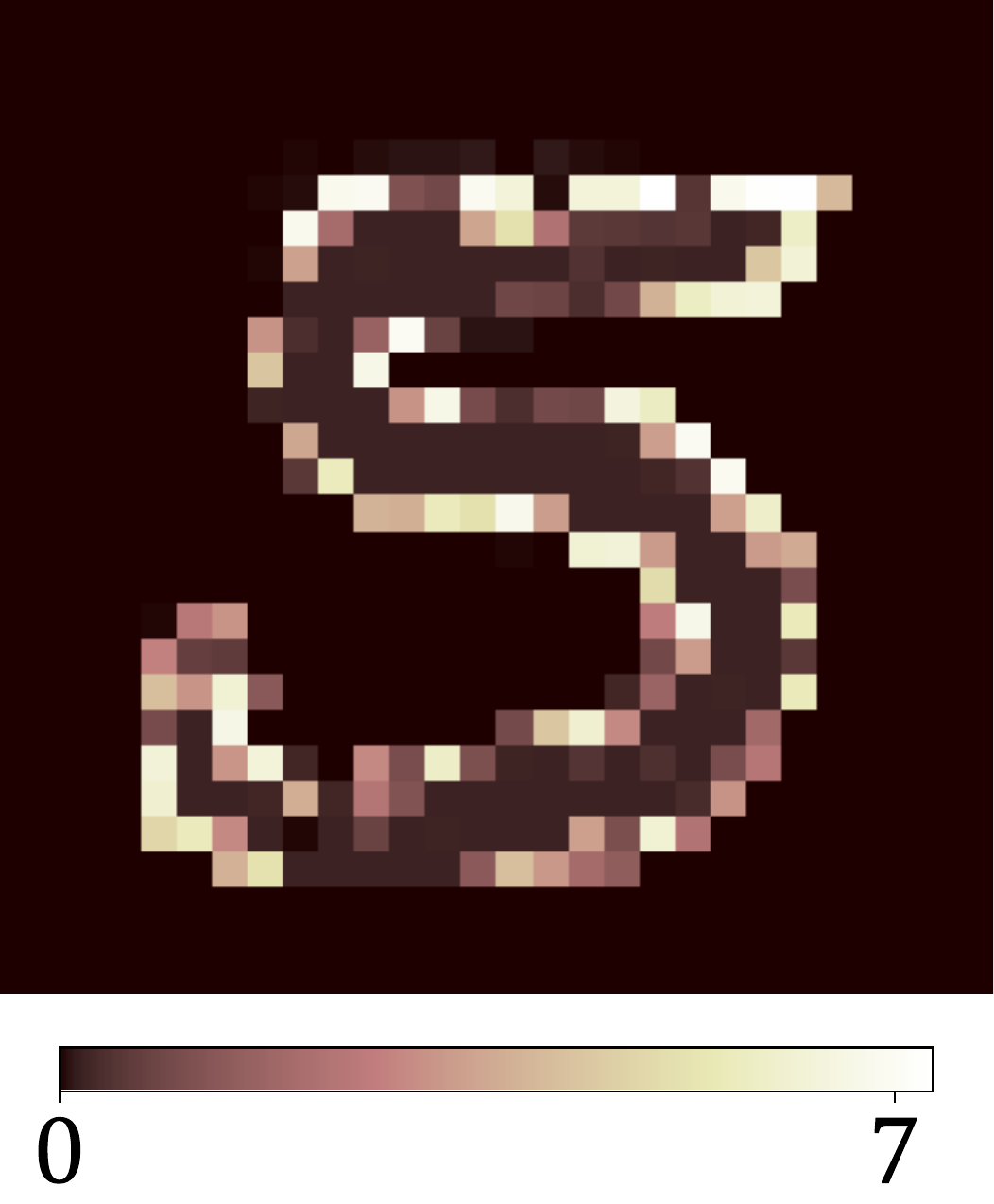} \end{minipage}
  \begin{minipage}[t]{0.095\textwidth} \centering \includegraphics[width=0.65in]{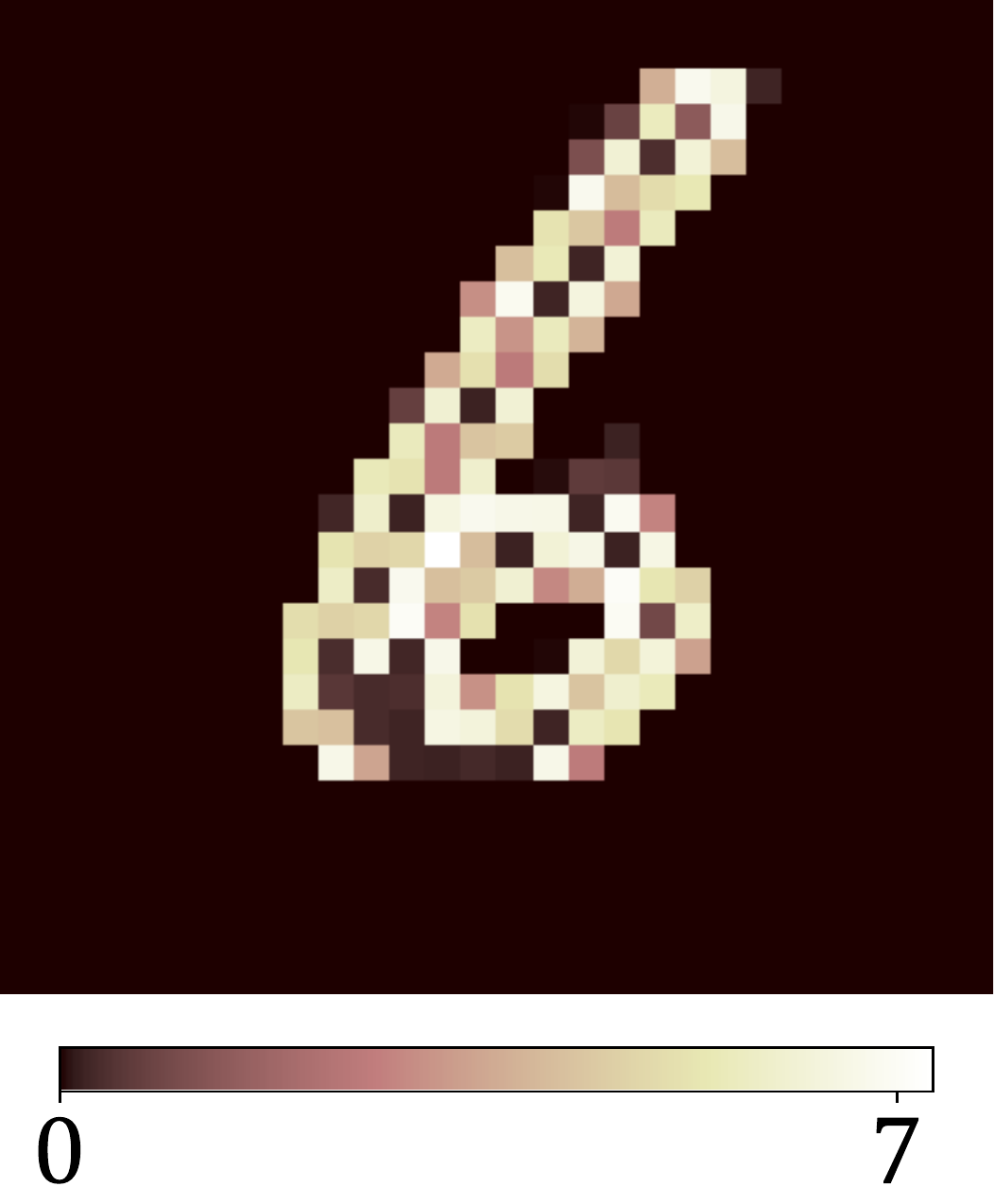} \end{minipage}
  \begin{minipage}[t]{0.095\textwidth} \centering \includegraphics[width=0.65in]{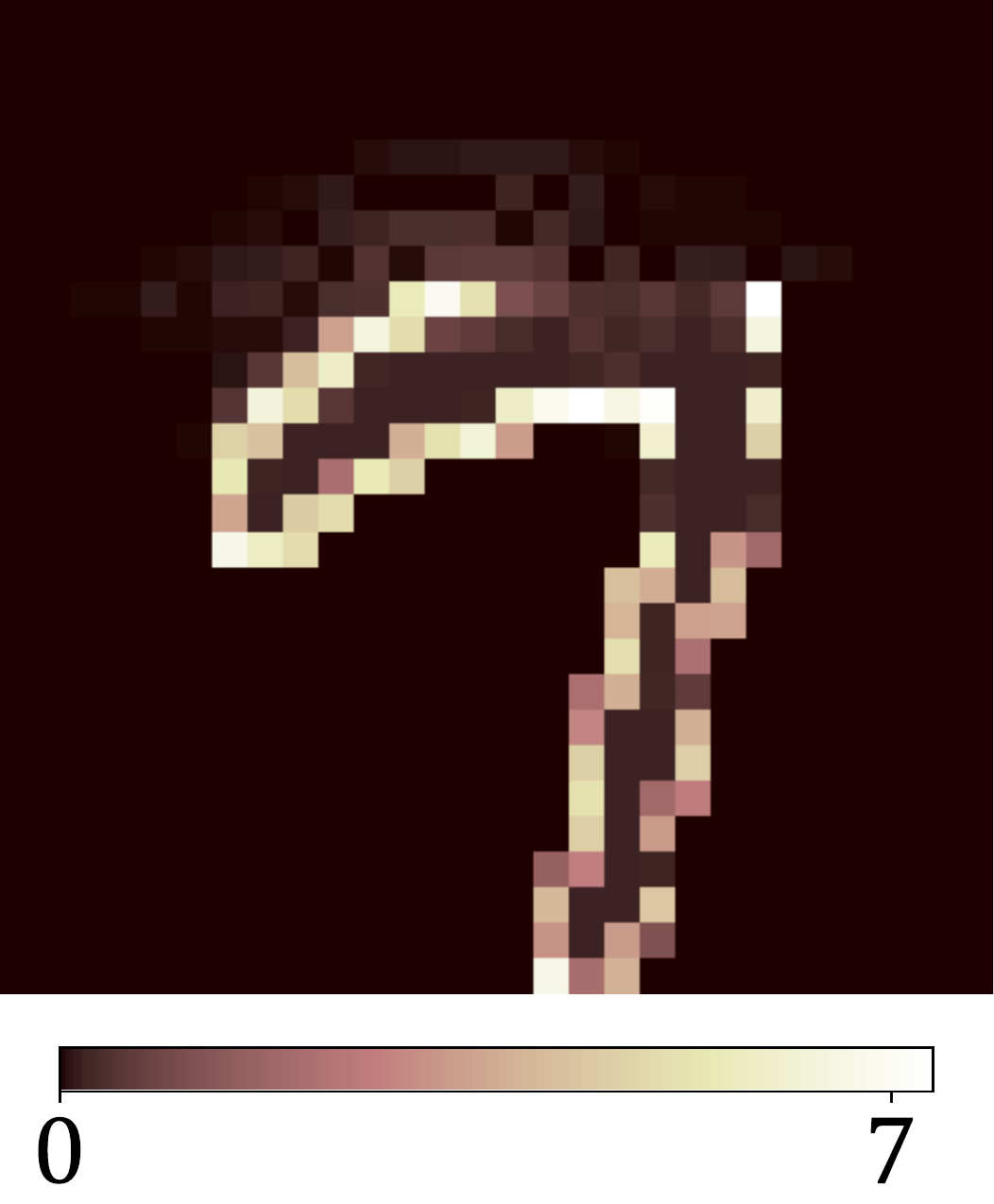} \end{minipage}
  \begin{minipage}[t]{0.095\textwidth} \centering \includegraphics[width=0.65in]{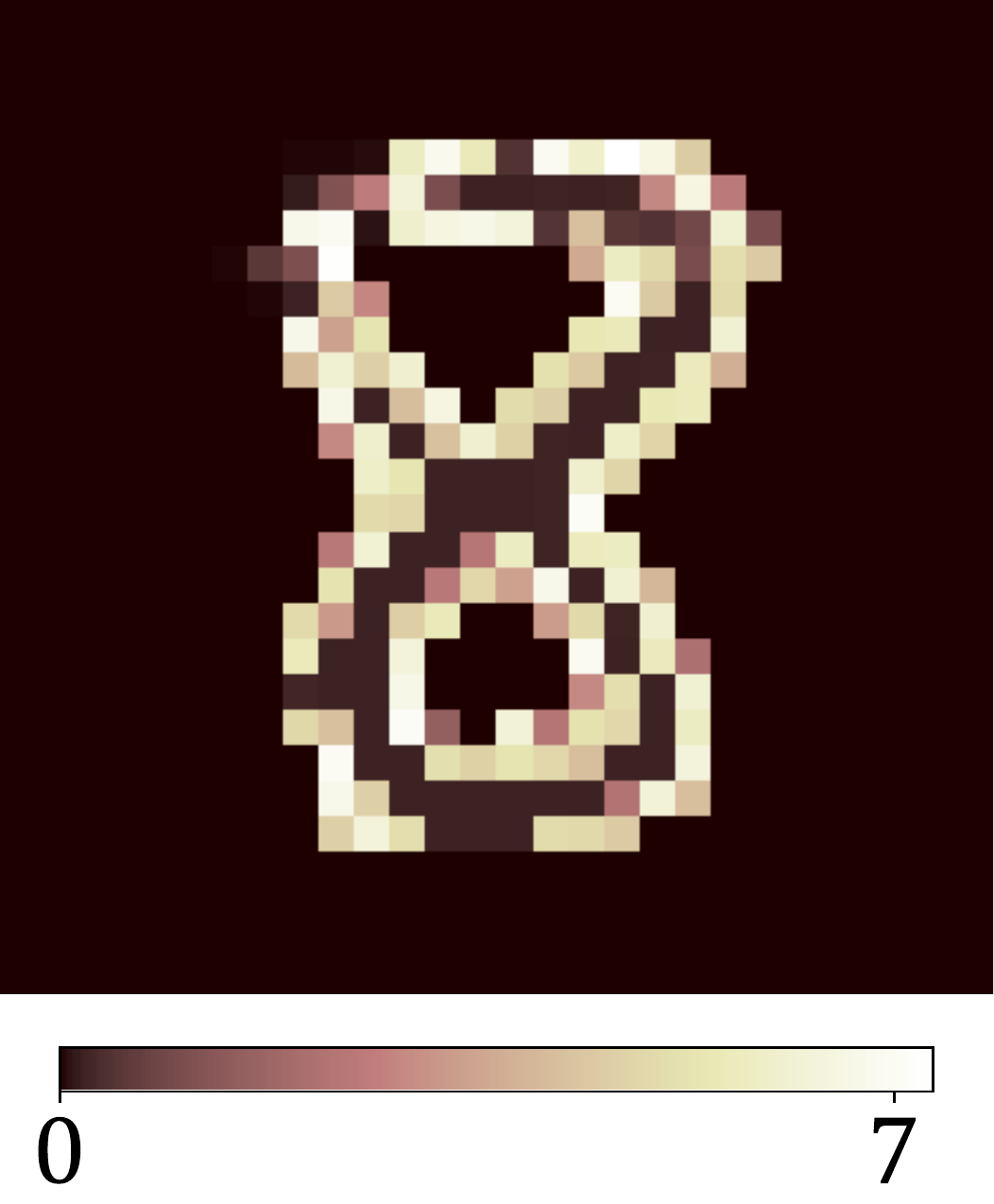} \end{minipage}
  \begin{minipage}[t]{0.095\textwidth} \centering \includegraphics[width=0.65in]{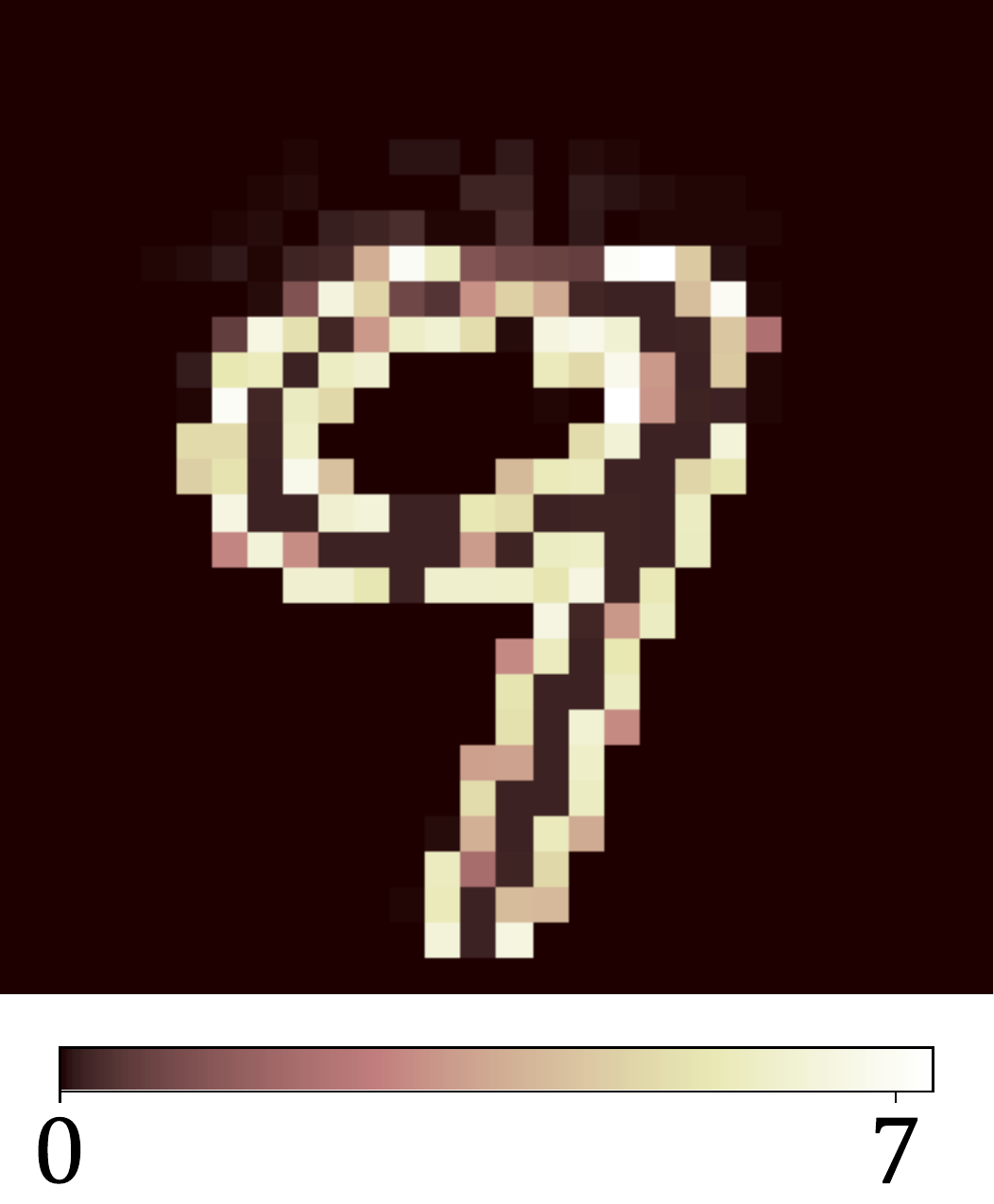} \end{minipage}
  \label{entropy subfigure}}
  \centering
  \subfigure[Amount of Embedded Information]{
  \begin{minipage}[t]{0.095\textwidth} \centering \includegraphics[width=0.65in]{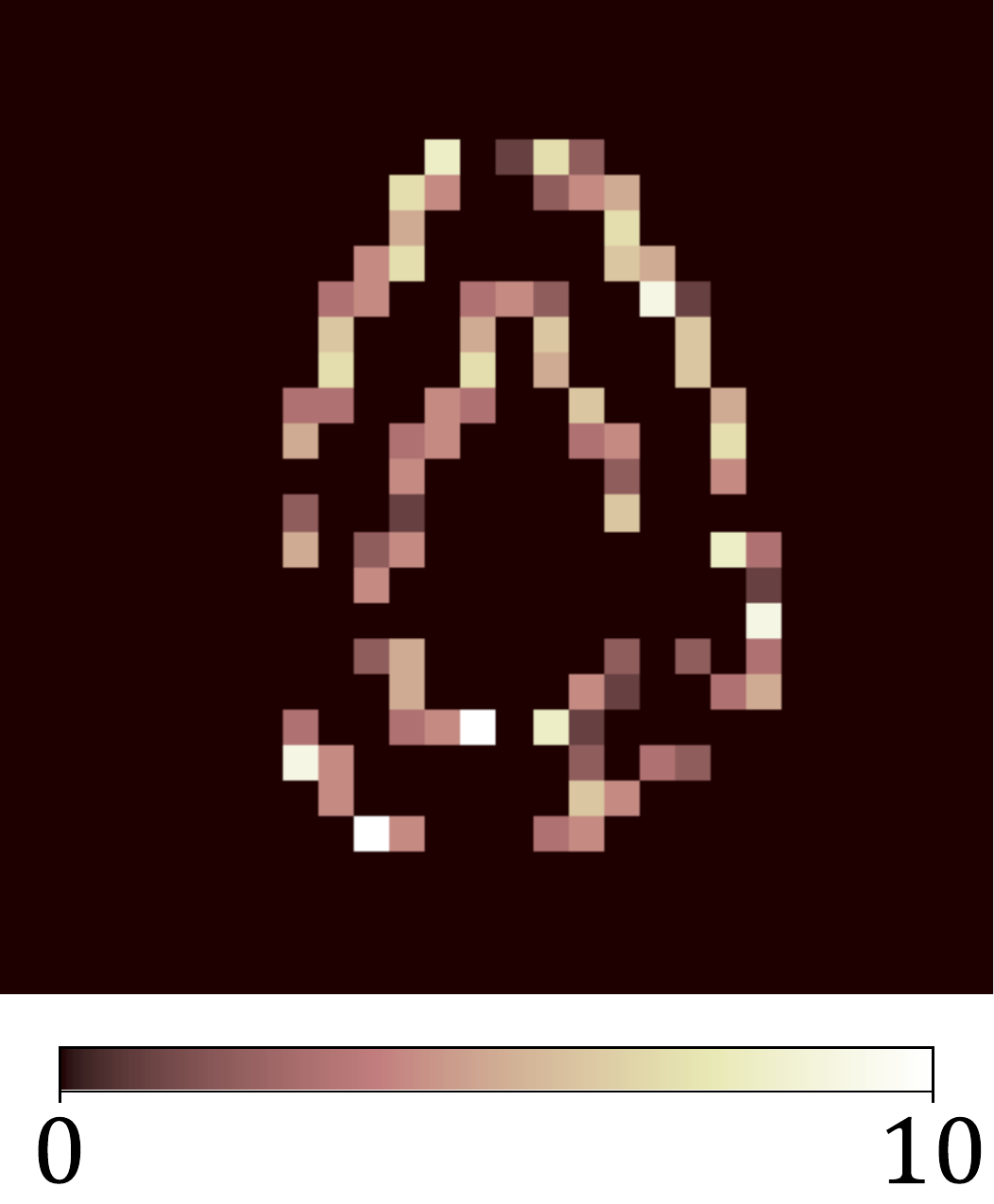} \end{minipage}
  \begin{minipage}[t]{0.095\textwidth} \centering \includegraphics[width=0.65in]{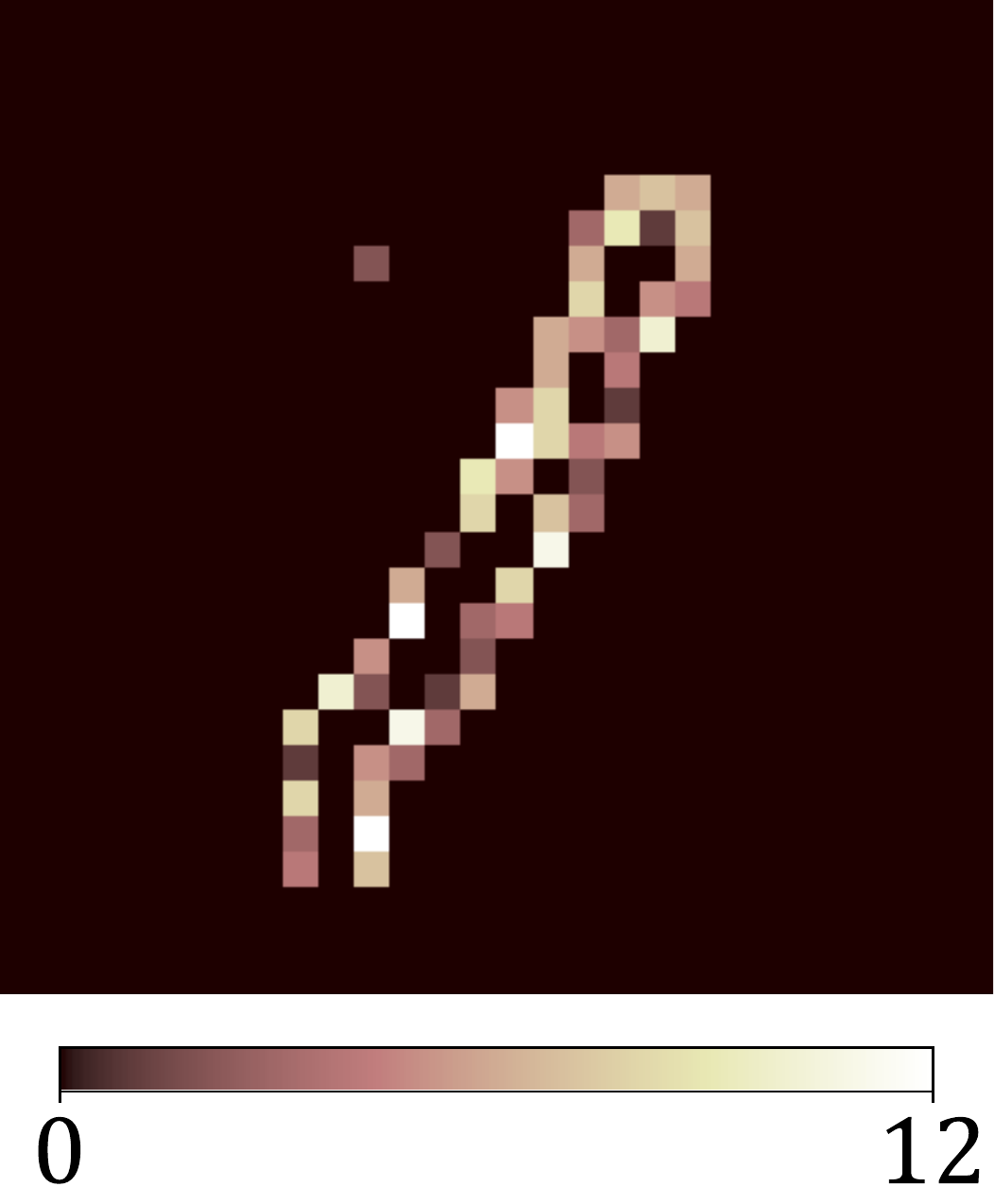} \end{minipage}
  \begin{minipage}[t]{0.095\textwidth} \centering \includegraphics[width=0.65in]{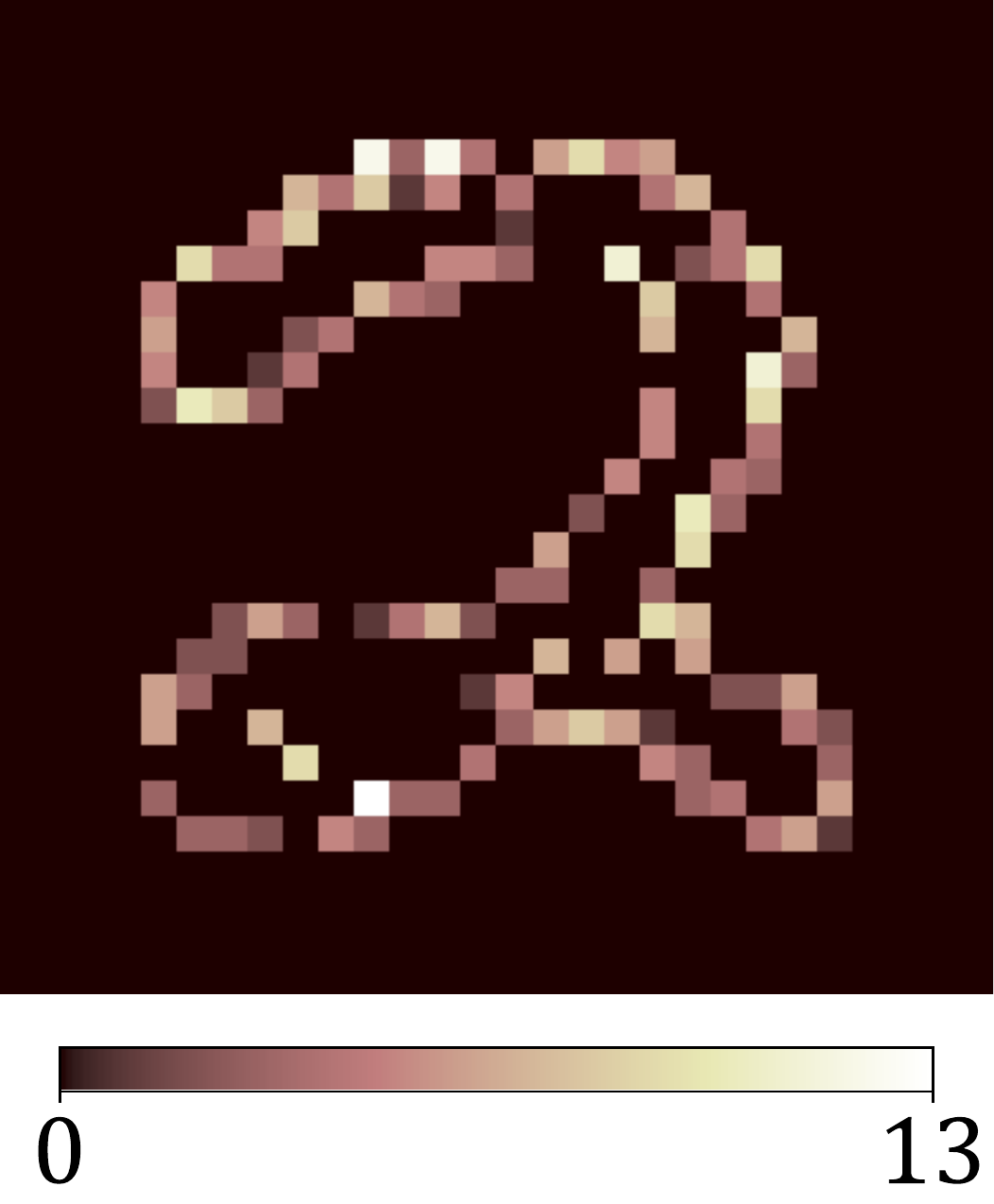} \end{minipage}
  \begin{minipage}[t]{0.095\textwidth} \centering \includegraphics[width=0.65in]{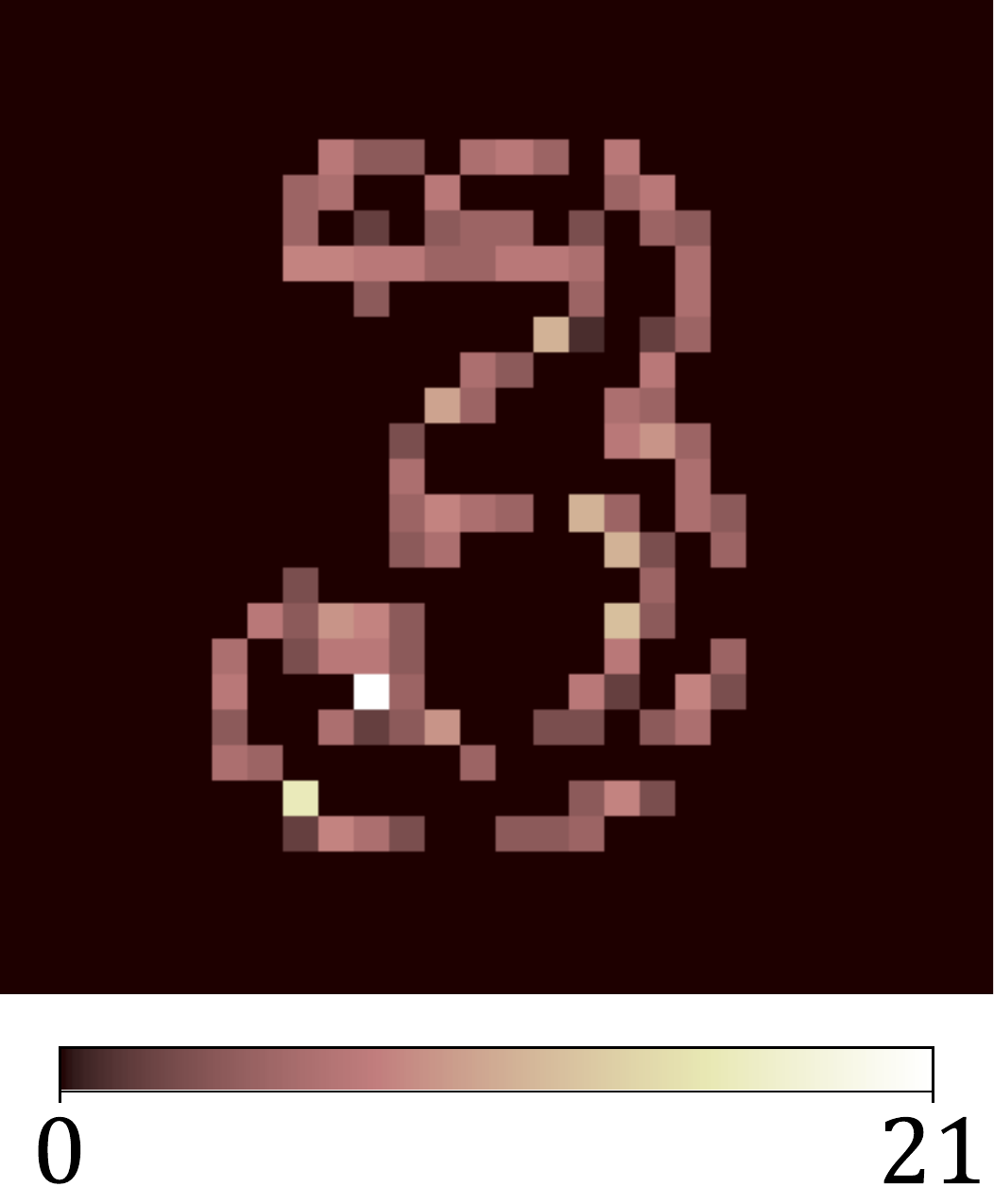} \end{minipage}
  \begin{minipage}[t]{0.095\textwidth} \centering \includegraphics[width=0.65in]{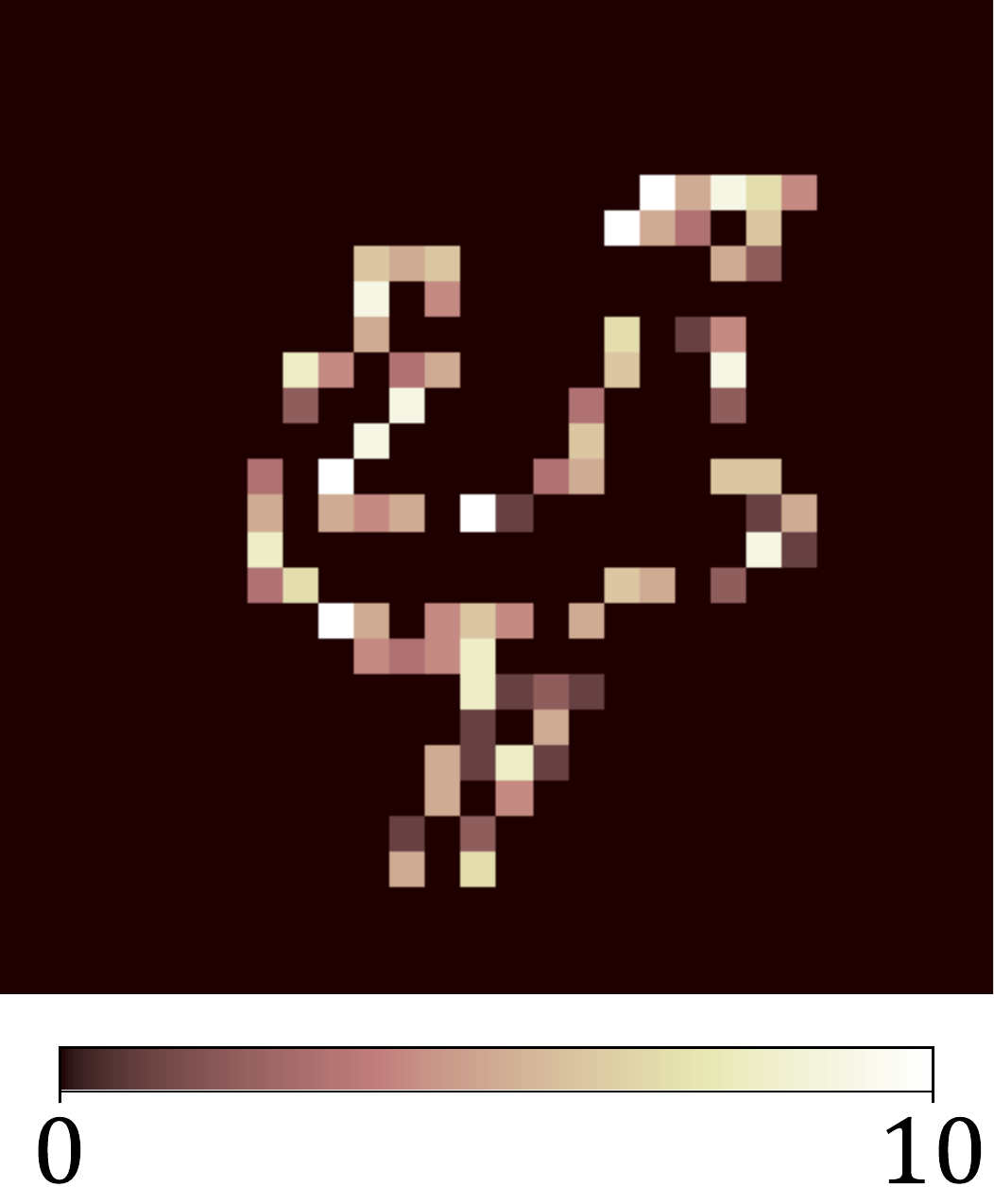} \end{minipage}
  \begin{minipage}[t]{0.095\textwidth} \centering \includegraphics[width=0.65in]{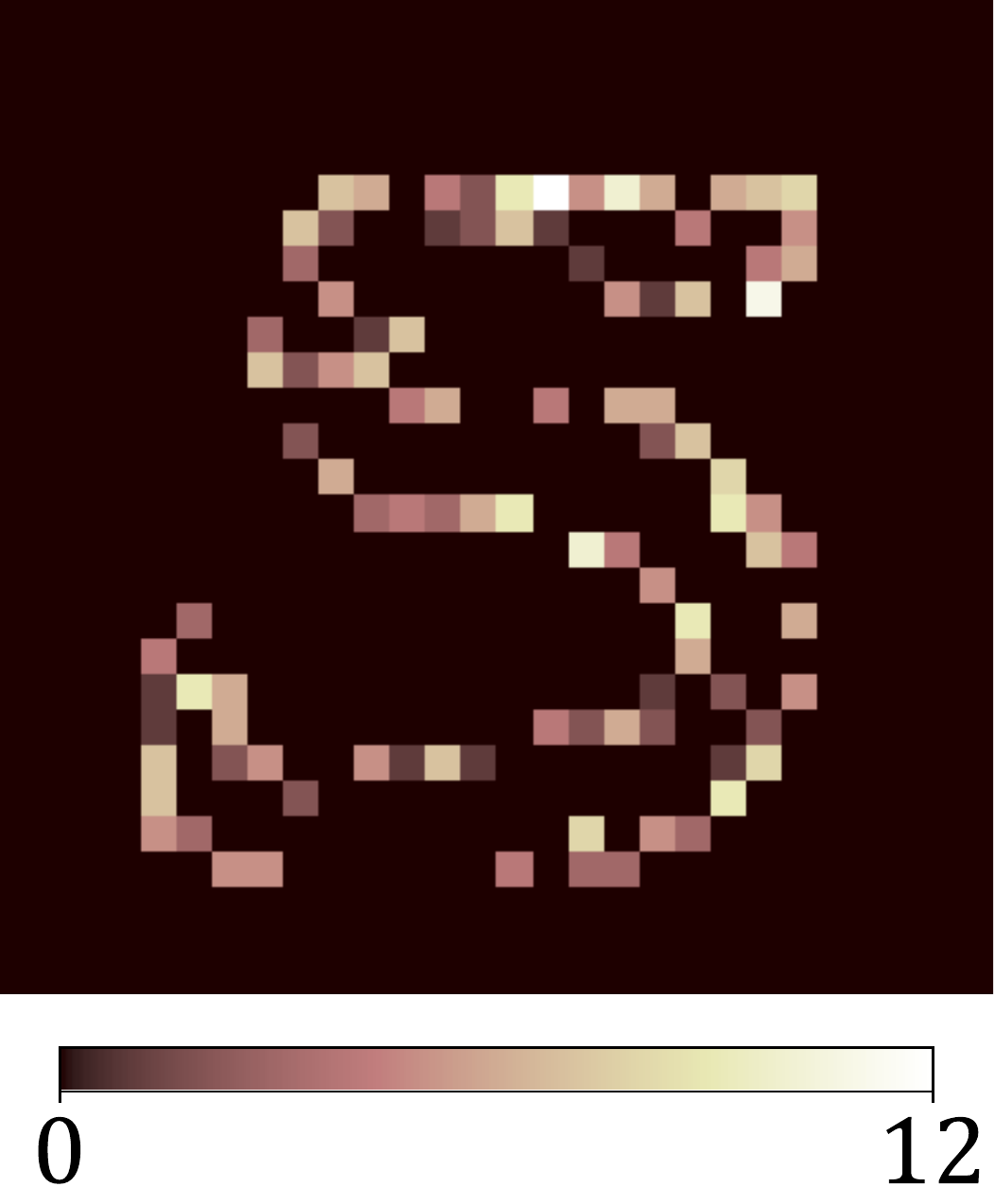} \end{minipage}
  \begin{minipage}[t]{0.095\textwidth} \centering \includegraphics[width=0.65in]{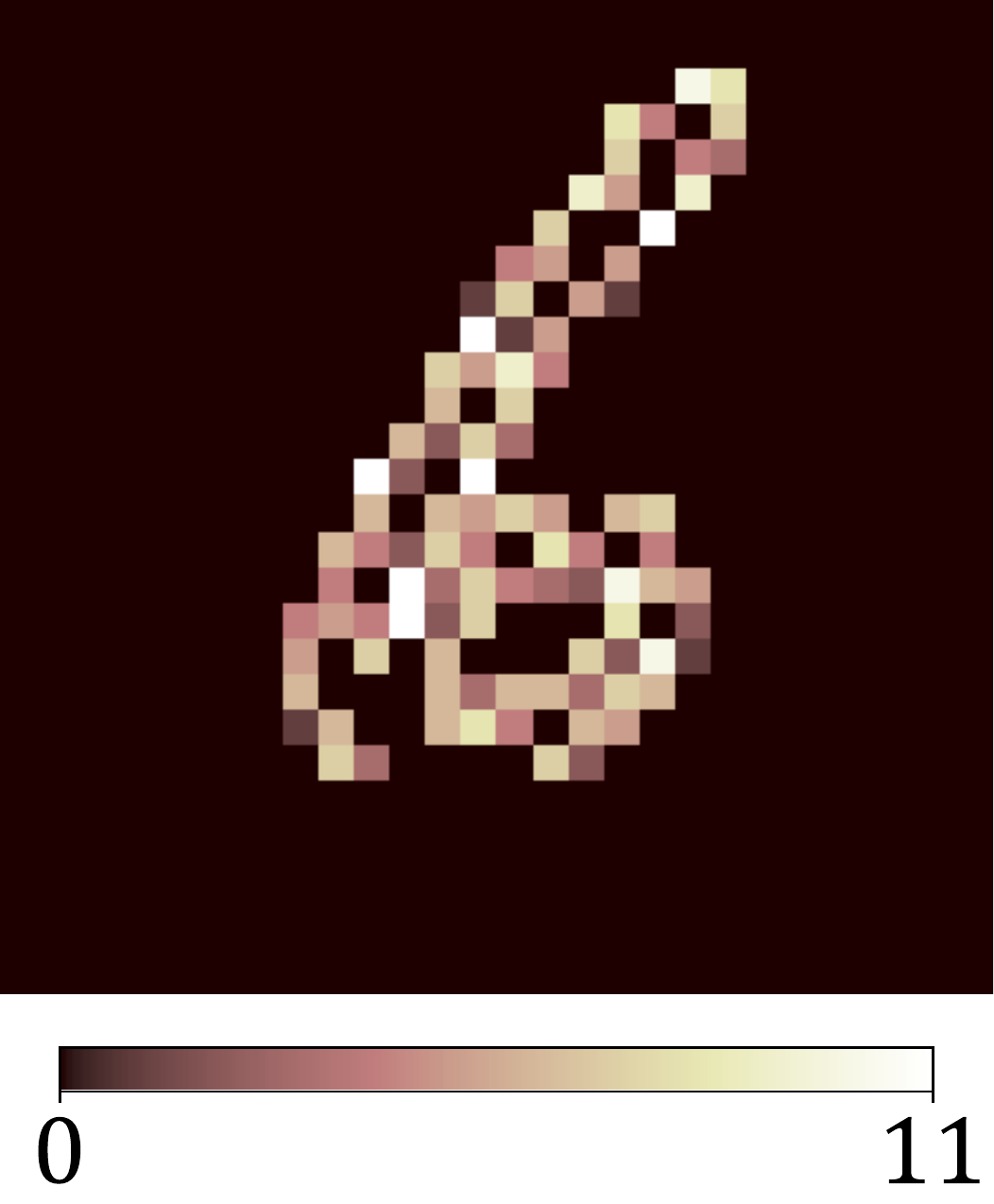} \end{minipage}
  \begin{minipage}[t]{0.095\textwidth} \centering \includegraphics[width=0.65in]{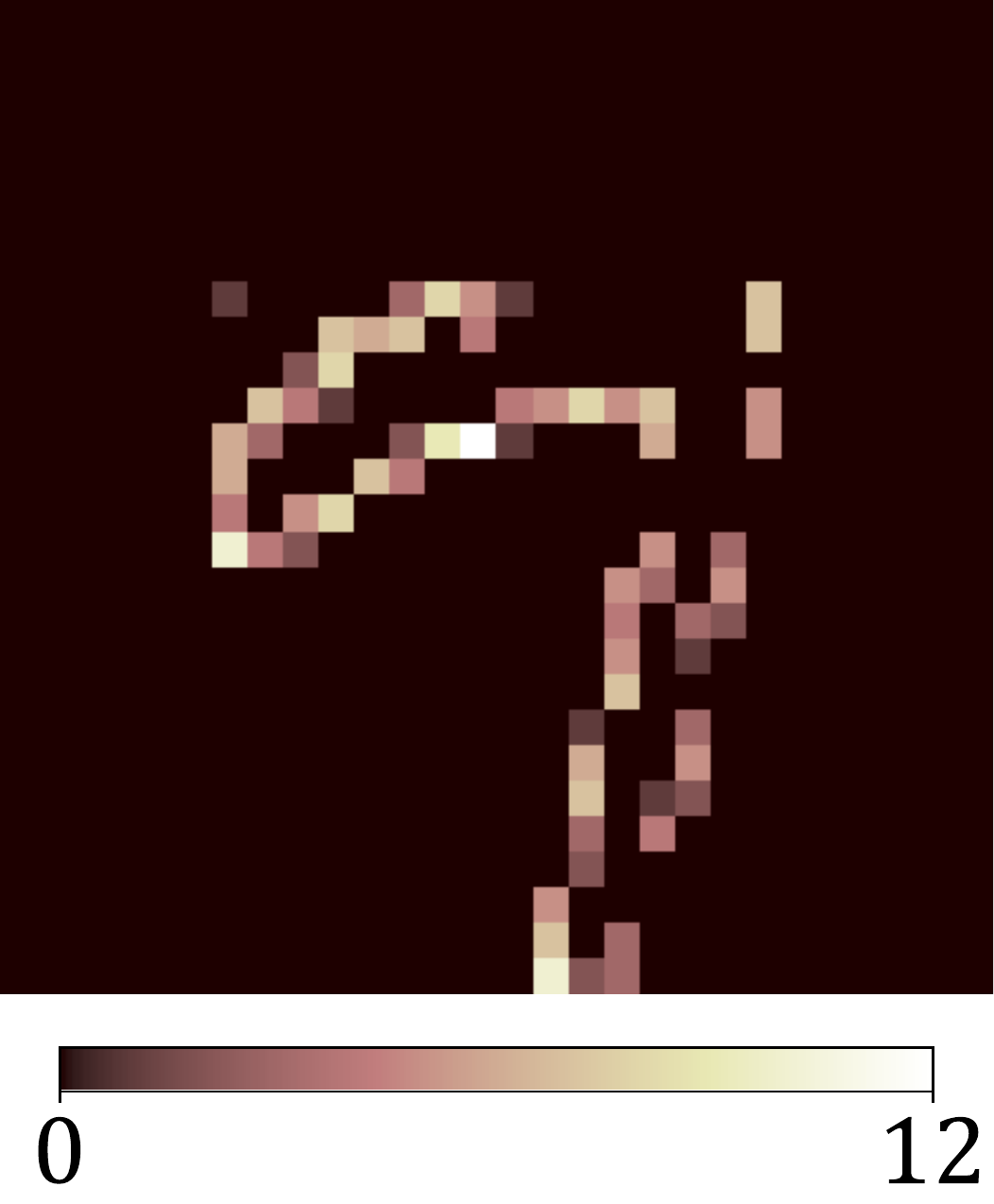} \end{minipage}
  \begin{minipage}[t]{0.095\textwidth} \centering \includegraphics[width=0.65in]{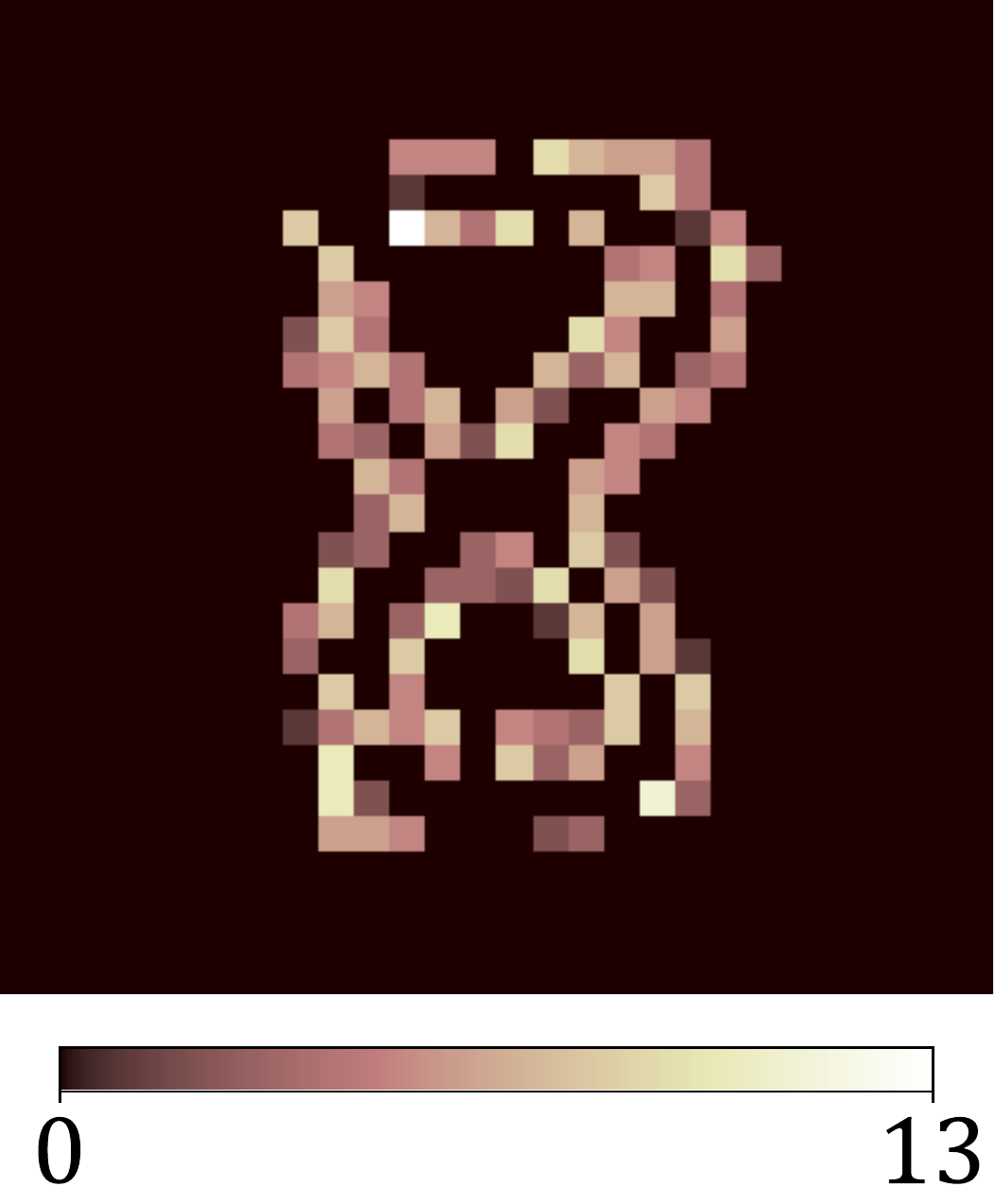} \end{minipage}
  \begin{minipage}[t]{0.095\textwidth} \centering \includegraphics[width=0.65in]{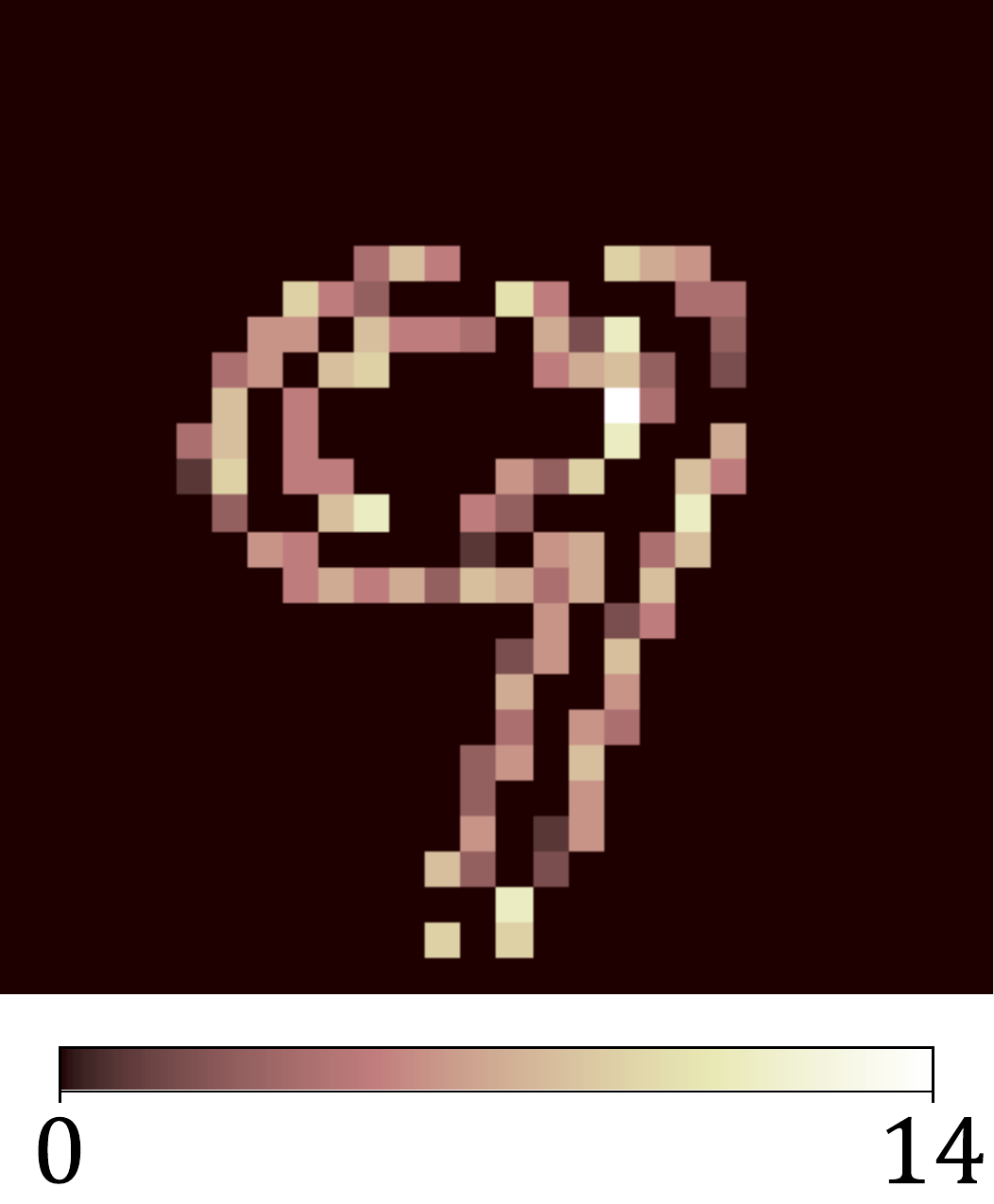} \end{minipage}
  \label{bit subfigure}}
  \caption{The entropy $H(p(x_i|x_{<i}))$ (bit) and the amount of embedded information (bit) of stego-images on MNIST, where the latter is highly dependent on the former. From an overall perspective, more information is embedded into pixels with large entropy adaptively and vice versa.}
  \label{entropy figure}
\end{figure*}

\section{Experiments} 
\label{section experiments}

\subsection{Setups}
\label{subsection setups}
We conducted experiments on MNIST \cite{lecun1998gradient}, Frey Faces\footnote{\url{http://www.cs.nyu.edu/~roweis/data/frey_rawface.mat}} and CIFAR-10 \cite{krizhevsky2009learning}. Details of the datasets are shown in Table \ref{dataset table}.

\begin{table}[h]
  \footnotesize
  \centering
  \caption{Dataset Statistics.}
  \label{dataset table}
  \begin{threeparttable}
  \begin{tabular}{m{45pt}<{\centering} m{38pt}<{\centering} m{31pt}<{\centering} m{31pt}<{\centering} m{31pt}<{\centering}}
    \toprule[1.5pt]
    \textbf{DATASET} 	& \textbf{SHAPE} & \textbf{MODE} & \textbf{TRAIN} & \textbf{TEST}	\\
    \midrule[1.3pt]
	      MNIST         & 28 $\times$ 28    & GRAY  & 60,000            & 10,000      \\ \specialrule{0em}{1pt}{1pt}
	      Frey Faces    & 28 $\times$ 20    & GRAY  & 1,685\tnote{*}   & 280\tnote{*} \\ \specialrule{0em}{1pt}{1pt}
	      CIFAR-10		& 32 $\times$ 32    & RGB   & 50,000            & 10,000      \\
    \bottomrule[1.5pt]
  \end{tabular}
  \begin{tablenotes}
     \item[*] We split training set and test set of Frey Faces in the ratio of 6 to 1.
   \end{tablenotes}
   \end{threeparttable}
\end{table}

We followed the original settings of PixelCNN++ in \cite{salimans2017pixelcnn++}. For CIFAR-10, we utilized a public Pytorch version  pre-trained model\footnote{\url{https://github.com/pclucas14/pixel-cnn-pp}}. For MNIST and Frey Faces, we trained the model from scratch and took the parameters with best test loss during 200-epoch training with batchsize 16 and learning rate 0.0002. As for stegosampling based on arithmetic coding, we set the fixed precision $prc$ to 26. We generated 5,000 cover-images, 5,000 stego-images by Pixel-Stega and 5,000 stego-image by method in \cite{yang2018provably} each dataset for further quantitative assessment. Some generated stego-images are illustrated  in Figure \ref{example figure}.

\subsection{Embedding Capacity}
\label{subsection embedding capacity}
In Table \ref{ER table}, we list the result of embedding capacity measured in embedding rate (ER). As we can see, autoregressive-model-based steganographic methods have much higher embedding capacity than other generative methods.
Compared with method in \cite{yang2018provably}, the result of Pixel-Stega on MNIST does not seem to show too much superiority. As discussed above, the amount of information that can be embedded at each time step for Pixel-Stega is actually effected by the entropy of pixels $H(p(x_i|x_{<i}))$. Images in MNIST have similar pattern of black background and white handwritten digits in the middle, which results in low entropy for most pixels. We visualize the entropy $H(p(x_i|x_{<i}))$ as well as the amount of embedded information of each pixel in Figure \ref{entropy figure}. We found that the uncertainty is mainly provided by pixels at the edge of the strokes. For most pixels in the background or inside the stroke, they have zero entropy and cannot contribute to any space to conceal secret messages.

\begin{table}[h]	
  \footnotesize
  \centering
  \caption{Experimental Results of Embedding Rate (bpp).}
  \label{ER table}
  \begin{threeparttable}
  \begin{tabular}{m{40pt}<{\centering} p{79pt}<{\centering} m{80pt}<{\centering}}
    \toprule[1.5pt]
    \textbf{MODEL} & \textbf{DATASET}  & \textbf{ER $\uparrow$} \\
    \midrule[1.3pt]
    \cite{liu2017coverless}	 							& MNIST		& $\approx$ 0.0030 	\\ \specialrule{0em}{1pt}{1pt}
    \cite{hu2018novel}									& CelebA \cite{liu2015faceattributes}/Food101\tablefootnote{\url{https://www.vision.ee.ethz.ch/datasets_extra/food-101/}}& $\approx$ 0.0733 	\\ \specialrule{0em}{1pt}{1pt}
    \cite{chen2018provably}							& Anime\tablefootnote{\url{https://drive.google.com/file/d/1yOrpEjEYU8LXl8h-k7gVaaQ0ZdGYfnU_/view?usp=sharing}}				& $\approx$ 0.0977 	\\ \specialrule{0em}{1pt}{1pt}
    \cite{li2020generative}								& CelebA \cite{liu2015faceattributes}				& $\approx$ 0.0733 	\\ \specialrule{0em}{1pt}{1pt}
    \hline \specialrule{0em}{1pt}{1pt}
    \multirow{3}{*}{\cite{yang2018provably}\tnote{*}}	& MNIST		    & \textbf{1.0000 $\pm$ 0.0000}	    \\ \specialrule{0em}{1pt}{1pt}
    											& Frey Faces    & 1.0000 $\pm$ 0.0000	            \\ \specialrule{0em}{1pt}{1pt}
    											& CIFAR-10	    & 1.0000 $\pm$ 0.0000	            \\ \specialrule{0em}{1pt}{1pt}
    \hline \specialrule{0em}{1pt}{1pt}
    \multirow{3}{*}{Pixel-Stega\tnote{*}}				& MNIST			& 0.5840 $\pm$ 0.1879				\\ \specialrule{0em}{1pt}{1pt}
    											& Frey Faces	& \textbf{4.0479 $\pm$ 0.1403}	    \\ \specialrule{0em}{1pt}{1pt}
    											& CIFAR-10	    & \textbf{4.3028 $\pm$ 0.8502}	    \\ \specialrule{0em}{1pt}{1pt}
    \bottomrule[1.5pt]
  \end{tabular}
  \begin{tablenotes}
     \item[*] Autoregressive-model-based method.
   \end{tablenotes}
   \end{threeparttable}
\end{table}

\begin{table*}[h]	
  \footnotesize
  \centering
  \caption{Experimental Results of Anti-steganalysis.}
  \label{steganalysis table}
  \begin{tabular}{m{34pt}<{\centering} m{40pt}<{\centering} | m{21pt}<{\centering} m{21pt}<{\centering} m{21pt}<{\centering} m{21pt}<{\centering} | m{21pt}<{\centering} m{21pt}<{\centering} m{21pt}<{\centering} m{21pt}<{\centering} | m{21pt}<{\centering} m{21pt}<{\centering} m{21pt}<{\centering} m{21pt}<{\centering}}
    \toprule[1.5pt]
    \multirow{2}{*}{\textbf{DATASET}} 	& \multirow{2}{*}{\textbf{MODEL}} & \multicolumn{4}{c|}{\textbf{XuNet} \cite{xu2016structural}} & \multicolumn{4}{c|}{\textbf{Yedroudj-Net}\cite{yedroudj2018yedroudj}} & \multicolumn{4}{c}{\textbf{SRNet} \cite{boroumand2018deep}} \\
    \cline{3-14} \specialrule{0em}{1pt}{1pt}
    && $Acc$ & $p$ & $r$ & $f_1$ & $Acc$ & $p$ & $r$ & $f_1$ & $Acc$ & $p$ & $r$ & $f_1$ \\
    \midrule[1.3pt]
    \multirow{2}{*}{MNIST}		
    & \cite{yang2018provably}	    & 1.0000 & 1.0000 & 1.0000 & 1.0000 & 1.0000 & 1.0000 & 1.0000 & 1.0000 & 1.0000 & 1.0000 & 1.0000 & 1.0000 	\\ \specialrule{0em}{1pt}{1pt}
    & Pixel-Stega					& 0.5278 & 0.5292 & 0.5028 & 0.5128 & 0.5215 & 0.5240 & 0.5344 & 0.5234 & 0.5023 & 0.5028 & 0.5680 & 0.5298   	\\ \hline \specialrule{0em}{2pt}{1pt}
    \multirow{2}{*}{Frey Faces}			
    & \cite{yang2018provably}	    & 0.5158 & 0.5162 & 0.5122 & 0.5118 & 0.5130 & 0.5128 & 0.6264 & 0.5462 & 0.5130 & 0.5118 & 0.5650 & 0.5327     \\ \specialrule{0em}{1pt}{1pt}
    	& Pixel-Stega				& 0.5081 & 0.5078 & 0.5178 & 0.5099 & 0.5257 & 0.5377 & 0.3976 & 0.4506 & 0.5551 & 0.5784 & 0.4212 & 0.4836     \\ \hline \specialrule{0em}{2pt}{1pt}
	\multirow{2}{*}{CIFAR-10}	
	& \cite{yang2018provably}	    & 0.5323 & 0.5348 & 0.4960 & 0.5131 & 0.5328 & 0.5385 & 0.4594 & 0.4942 & 0.5572 & 0.5900 & 0.3932 & 0.4631     \\ \specialrule{0em}{1pt}{1pt}
    	& Pixel-Stega				& 0.5294 & 0.5332 & 0.4990 & 0.5094 & 0.5561 & 0.5711 & 0.4613 & 0.5061 & 0.5640 & 0.5793 & 0.4717 & 0.5183     \\
    \bottomrule[1.5pt]
  \end{tabular}
\end{table*}

On the contrary, results on Frey Faces and CIFAR-10 illustrate that Pixel-Stega has a prominent advantage in embedding capacity than method in \cite{yang2018provably}. Images in these datasets are more diverse, which means their pixels have large entropy. The results reveal that Pixel-Stega is able to make the full use of $H(p(x_i|x_{<i}))$ to its potential. However, method in \cite{yang2018provably} only encodes secret messages to the LSB of pixels, which is distinctly a waste of the mount of information that $p(x_i|x_{<i})$ supplies (only 1 bit information is used).

\subsection{Imperceptibility}
\label{subsection imperceptibility}
We measured the distortion between the equivalent distribution $q(x_i|x_{<i})$ and the original conditional distribution $p(x_i|x_{<i})$ by Kullback-Leibler divergence (KLD) and Jensen–Shannon divergence (JSD).

\begin{table}[h]
  \footnotesize
  \centering
  \caption{Experimental Results of $D_{KL}(q||p)$ and $D_{JS}(q||p)$ (bit).}
  \label{quality table}
  \begin{tabular}{m{37pt}<{\centering} p{45pt}<{\centering} m{53pt}<{\centering} m{53pt}<{\centering}}
    \toprule[1.5pt]
    \textbf{DATASET} 	& \textbf{MODEL} & \textbf{KLD $\downarrow$} & \textbf{JSD $\downarrow$}	\\
    \midrule[1.3pt]
    \multirow{2}{*}{MNIST}		& \cite{yang2018provably}	& 1.2480E+01		    & 1.6928E-01            \\ \specialrule{0em}{1pt}{1pt}
    							& Pixel-Stega				& \textbf{6.2003E-06}   & \textbf{2.0988E-06}   \\ \hline \specialrule{0em}{2pt}{1pt}
    \multirow{2}{*}{Frey Faces}	& \cite{yang2018provably}	& 2.0547E-02			& 3.0439E-03            \\ \specialrule{0em}{1pt}{1pt}
    							& Pixel-Stega				& \textbf{3.3624E-06}	& \textbf{9.6097E-07}   \\ \hline \specialrule{0em}{2pt}{1pt}
	\multirow{2}{*}{CIFAR-10}	& \cite{yang2018provably}	& 1.2132E-02 			& 2.3270E-03            \\ \specialrule{0em}{1pt}{1pt}
    							& Pixel-Stega				& \textbf{1.2151E-05}	& \textbf{3.9509E-06}   \\
    \bottomrule[1.5pt]
  \end{tabular}
\end{table}

According to the results listed in Table \ref{quality table}, the distortion between $p$ and $q$ of method in \cite{yang2018provably} can be large, especially on MNIST. As discussed, due to the simplex pattern of images in MNIST, the probability mass of most pixels almost concentrates on a single value thus leading to an extremely sharp distribution. Under such circumstances, using rejection sampling strategy to embed secret messages into the LSB of pixels is not reasonable. As a result, it fails to generate realistic stego-images (Figure \ref{rejection mnist figure}). In contrast, Pixel-Stega has more negligible distortion, which provides evidence that Pixel-Stega is adaptive to all kinds of distribution $p(x_i|x_{<i})$, no matter sharp or flat.

\begin{table}[h]	
  \footnotesize
  \centering
  \caption{Experimental Results of Effective Embedding Rate (bpp).}
  \label{EER table}
  \begin{tabular}{m{31pt}<{\centering} p{40pt}<{\centering} m{35pt}<{\centering} m{35pt}<{\centering} m{35pt}<{\centering}}
    \toprule[1.5pt]
    \textbf{DATASET} 	& \textbf{MODEL} & \textbf{EER$_1 \uparrow$} & \textbf{EER$_2 \uparrow$} & \textbf{EER$_3 \uparrow$} \\
    \midrule[1.3pt]
    \multirow{2}{*}{MNIST}		& \cite{yang2018provably}       & 0.0000            & 0.0000            & 0.0000 	        \\ \specialrule{0em}{1pt}{1pt}
    							& Pixel-Stega					& \textbf{0.5515}   & \textbf{0.5589}   & \textbf{0.5813}	\\ \hline \specialrule{0em}{2pt}{1pt}
    \multirow{2}{*}{Frey Faces}	& \cite{yang2018provably}       & 0.9684            & 0.9740            & 0.9740 	        \\ \specialrule{0em}{1pt}{1pt}
    							& Pixel-Stega					& \textbf{3.9824}   & \textbf{3.8399}   & \textbf{3.6019}   \\ \hline \specialrule{0em}{2pt}{1pt}
	\multirow{2}{*}{CIFAR-10}	& \cite{yang2018provably}       & 0.9344            & 0.8856            & 0.9354	        \\ \specialrule{0em}{1pt}{1pt}
    							& Pixel-Stega					& \textbf{4.0500}   & \textbf{3.8196}   & \textbf{3.7524}	\\
    \bottomrule[1.5pt]
  \end{tabular}
\end{table}

To further evaluate the imperceptibility, we tested the anti-steganalysis ability by three state-of-the-art steganalysis approaches, XuNet \cite{xu2016structural}, Yedroudj-Net \cite{yedroudj2018yedroudj} and SRNet \cite{boroumand2018deep}. Small adjustments were made to let the networks fit the shape of stego-images. We used 10-fold cross-validation to distinguish 5,000 cover-images and 5,000 stego-images. Each model was trained for 50 epochs with batchsize 64 and learning rate 0.01.

According to the results in Table \ref{steganalysis table}, we found Pixel-Stega achieves nearly perfect imperceptibility (closer to 0.5 is better), which further indicates its adaptability to all kinds of output distribution $p(x_i|x_{<i})$. Despite on MNIST, method in \cite{yang2018provably} seems also qualified to resist steganalysis. However, it exposes secret messages directly to the monitoring party (the LSB of pixels is actually the secret message). On the contrary, Pixel-Stega maintains a higher level of security, as secret messages are really and truly hidden inside the stego-images. No meaningful information can be extracted even if some abnormity is noticed.


\subsection{Comprehensive Assessment}
\label{subsection EER}
To measure the performance of Pixel-Stega comprehensively, we calculated effective embedding rate (EER) \cite{zhang2021provably} with the corresponding accuracy result of steganalysis method in Table \ref{steganalysis table}. The result reveals that Pixel-Stega achieves higher comprehensive performance than method in \cite{yang2018provably}. Besides, since EER is always less than or equal to ER, Pixel-Stega also surpasses other previous generative image steganographic methods. According to Figure \ref{entropy figure}, Pixel-Stega embeds more information to pixels with larger entropy and vice versa, thereby both ensuring high embedding capacity and nearly perfect imperceptibility, so as to achieve satisfying comprehensive performance.

\section{Conclusion}
\label{section conclusion}
In this letter, we proposed Pixel-Stega, a generative image steganographic method based on autoregressive models and arithmetic coding. 
We found Pixel-Stega is able to embed secret messages adaptively on the basis of the entropy $H(p(x_i|x_{<i}))$ of pixels. More information is embedded into high entropy region and less into low entropy region, which means
(1) the explicit distribution $p(x_i|x_{<i})$ is used to its full potential;
(2) the stegosampling strategy is adaptive to either sharp or flat distributions.
As a result, Pixel-Stega achieved both high embedding capacity (up to 4.3 bpp) and nearly perfect imperceptibility (about 50\% detection accuracy) at the same time.
It is hoped that this letter can facilitate the development of generative image steganography.


%




\ifCLASSOPTIONcaptionsoff
  \newpage
\fi



%



\bibliographystyle{IEEEtran}
\bibliography{bare_jrnl}

%








\end{document}